\documentclass[10pt,journal,compsoc]{IEEEtran}
\usepackage{array}
\usepackage{multirow}
\usepackage{makecell}
\usepackage{adjustbox}
\usepackage{subfig}
\usepackage{graphicx}
\usepackage{amsmath,amssymb,amsthm}
\usepackage{xcolor}
\usepackage{color, soul}
\usepackage{stfloats}
\usepackage{amsfonts,amssymb}
\usepackage[justification=centering]{caption}
\usepackage{algorithm}
\usepackage{algorithmic}
\newcommand{\R}{{\mathbb R}}

\DeclareMathOperator{\Tr}{tr}

\def \calO {\mathcal{O}}
\newcommand{\psd}[2] {\mathcal{S}^+(#1,#2)}
\newcommand{\frmat}[2] {\mathbb{R}_*^{#1 \times #2}}

\def\etal{\emph{et al.}}

\def\ie{\emph{i.e.}}

\newtheorem{theorem}{Theorem}

\ifCLASSOPTIONcompsoc
\usepackage[nocompress]{cite}
\else
\usepackage{cite}
\fi
\ifCLASSINFOpdf
\else
\fi
\begin{document}
		%
		\title{Automatic Estimation of Self-Reported Pain by Trajectory Analysis in the Manifold of Fixed Rank Positive Semi-Definite Matrices}
		%
		%
		%
		%
		
	\author{Benjamin~Szczapa,~\IEEEmembership{Member,~IEEE,}
			Mohamed~Daoudi,~\IEEEmembership{Senior,~IEEE,}
			Stefano~Berretti,~\IEEEmembership{Senior,~IEEE,}
			Pietro~Pala,~\IEEEmembership{Senior,~IEEE,}
			Alberto~Del Bimbo,~\IEEEmembership{Senior,~IEEE,}
			and~Zakia~Hammal,~\IEEEmembership{Member,~IEEE}
			\IEEEcompsocitemizethanks{\IEEEcompsocthanksitem B. Szczapa is with the Univ. Lille, CNRS, Centrale Lille, UMR 9189 CRIStAL, F-59000 Lille, France \protect\\
				E-mail: benjamin.szczapa@univ-lille.fr
				\IEEEcompsocthanksitem M. Daoudi is with IMT Nord Europe, Institut Mines-Télécom, Univ. Lille, Centre for Digital Systems, F-59000 Lille, France, and Univ. Lille, CNRS, Centrale Lille, Institut Mines-Télécom, UMR 9189 CRIStAL, F-59000 Lille, France, E-mail: mohamed.daoudi@imt-nord-europe.fr
				\IEEEcompsocthanksitem S. Berretti, P. Pala and A. Del Bimbo are with the Department of Information Engineering, University of Florence, Italy
				\IEEEcompsocthanksitem Z. Hammal is with the Robotics Institute, Carnegie Mellon University, Pittsburgh, PA, USA}
		}

	\IEEEtitleabstractindextext{%
		\begin{abstract}
			We propose an automatic method to estimate self-reported pain based on facial landmarks extracted from videos. For each video sequence, we decompose the face into four different regions and the pain intensity is measured by modeling the dynamics of facial movement using the landmarks of these regions. A formulation based on Gram matrices is used for representing the trajectory of landmarks on the Riemannian manifold of symmetric positive semi-definite matrices of fixed rank. A curve fitting algorithm is used to smooth the trajectories and temporal alignment is performed to compute the similarity between the trajectories on the manifold. A Support Vector Regression classifier is then trained to encode extracted trajectories into pain intensity levels consistent with self-reported pain intensity measurement. Finally, a late fusion of the estimation for each region is performed to obtain the final predicted pain level. The proposed approach is evaluated on two publicly available datasets, the UNBCMcMaster Shoulder Pain Archive and the Biovid Heat Pain dataset. We compared our method to the state-of-the-art on both datasets using different testing protocols, showing the competitiveness of the proposed approach.
		\end{abstract}
		
		}

	\maketitle

	\IEEEdisplaynontitleabstractindextext

	%
	\IEEEpeerreviewmaketitle

	\IEEEraisesectionheading{\section{Introduction}}
	\IEEEPARstart{P}{ain} is an unpleasant sensory and emotional experience associated with actual or potential tissue damage caused by illness or injury~\cite{Merskey:1979}. 
	Pain assessment is necessary for differential diagnosis, choosing, monitoring, and evaluating treatment efficiency. 
	The assessment of pain is accomplished primarily through subjective self-reports using different medical scales like the Visual Analog Scale (VAS)---the most commonly used scale in clinical assessment~\cite{Aicher:2012, Farrar:2000, Jensen:2003, Jensen:2005}---or the Numerical Rating Scale (NRS)~\cite{Younger2009PainOA}. 
	However, while useful, self-reported pain is difficult to interpret due to subjectivity and personal experiences, and may be impaired or, in some circumstances, not even possible to obtain, such as for children, cognitively impaired patients or patients requiring breathing assistance~\cite{Hammaletal2018}. 
	To improve assessment of pain and guide treatment, objective measurement of pain from nonverbal behavior (\ie, facial expressions, head and body movements, and vocalizations) is emerging as a powerful option~\cite{Hammaletal2018, Werner:2019}.
	
	Extensive behavioral research has documented reliable facial indicators of pain~\cite{Prkachin:2008, Craig:2011, Kunz:2007, Prkachin:1992}. The core facial movements that have been found to discriminate the presence from the absence of pain are brow lowering, orbit tightening, upper-lip raising, nose wrinkling, and eye closure~\cite{Prkachin:2008}. Based on these findings, most efforts in automatic assessment of pain have focused on facial expression. Using either the Facial Action Coding System (FACS)~\cite{Prkachin:2008} or the holistic dynamics of the face, computational models have been trained to learn the association between discriminative facial features and pain occurrence or intensity~\cite{Hammaletal2018, Werner:2019}.

Building upon previous efforts, our primary measure for computational pain assessment is the dynamics of pain related facial movements~\cite{Prkachin:2008}. To assess the contribution of different facial regions in distinguishing between scores of pain level, facial movement dynamics were measured holistically (\ie, using the whole face) and by face-specific regions (\ie, eyes, eyebrows, mouth, and nose, separately and in combination). To capture changes in the dynamics of facial movement relevant to pain expression, we propose an original framework based on the temporal evolution of facial landmarks modeled as a trajectory on a Riemannian manifold. This formulation has shown promising results in action recognition~\cite{Devanne2015Cybernetics, amor2016action, Daoudi19ACII, Szczapa2019ICCVW} and in facial expression recognition~\cite{Szczapa2019ICCVW, KacemDABP20}. 
In our case, Gram matrices are computed from facial landmarks at each video frame and their temporal evolution is modeled as a trajectory on the Riemannian manifold of symmetric positive semi-definite (PSD) matrices~\cite{Szczapa2019ICCVW}. With this representation, pain estimation is modeled as the problem of computing similarity between trajectories on the manifold, then using a Support Vector Regression~\cite{DruckerBKSV96SVR} model to predict pain scores. 

In summary, the main contributions of this work are:
\begin{itemize}
	\item We propose a solution to model the temporal dynamics of facial landmarks position and velocity as Gram matrix trajectories on the Positive Semi-Definite (PSD) manifold; 
	\item Using Gram matrix trajectories, we estimate pain score at sequence-level, rather than at frame-level. 
	\item We present state-of-the-art results in VAS score estimation on two benchmark datasets.
\end{itemize}

The remaining of the paper is organized as follows: In Section~\ref{sect:related-work}, we summarize previous approaches to pain detection that are more relevant to the proposed model; 
In Section~\ref{sect:face-representation}, we introduce the face representation based on facial landmarks, and in Section~\ref{sect:face-dynamics} we extend it to include the temporal dynamics. 
The way we use the above representation for pain estimation is illustrated in Section~\ref{sect:pain-estimation}. Experimental results on two pain benchmark datasets are reported in Section~\ref{sect:results}. Finally, in Section~\ref{sect:conclusions}, we present a critical discussion of our work with perspectives for future development.

\section{Related Work}\label{sect:related-work}

%
%
Significant efforts have been made in human behavioral studies to identify reliable and valid facial indicators of pain~\cite{Prkachin:2008, Craig:2011, Kunz:2007, Prkachin:1992}. In these studies, pain expression and intensity were characterized at the frame level by highly trained human coders that annotated anatomical facial actions using the Facial Action Coding System (FACS)~\cite{Ekman:2002}. However, manual FACS based pain assessment requires over a hundred hours of training for FACS certification, and approximately an hour or more to manually annotate a minute of video. The intensive time required to annotate videos using the manual FACS makes it ill suited for daily clinical use. This limitation led to the emergence of considerable efforts in computer vision and machine learning for automatic pain assessment of self-reported pain (\ie, VAS based measurement) and observed pain (\ie, FACS based), respectively~\cite{Hammaletal2018}. 
Since the goal of our work is to automatically assess the self-reported pain, the state-of-the-art on FACS based pain assessment is not reviewed here (see~\cite{Hammaletal2018} for a detailed review).  

Using the UNBC-McMaster Shoulder Pain Archive~\cite{UNBC-McMaster} a few recent efforts have investigated video based measurement of self-reported Visual Analog Scale (VAS) pain intensity scores. The VAS is a self-reported pain score that indicates on a 0 to 10 scale the intensity of pain (where 0 corresponds to no pain, and 10 to the worst possible pain). 
For instance, Martinez~\etal~\cite{Martinez:cvrpw17} proposed a two-step learning approach to estimate pain scores consistent with the self-reported VAS. The authors employed a Recurrent Neural Network (RNN) to first estimate the Prkachin and Solomon Pain Intensity score (PSPI) at frame-level from face images. The estimated PSPI scores were then fed into a personalized Hidden Conditional Random Fields (HCRF) to derive a pain score consistent with the VAS. 
Liu~\etal~\cite{LiuPRP17-DeepFaceLIFT} proposed a two-stage personalized model, named DeepFaceLIFT, for automatic estimation of the self-reported VAS pain score. This approach is based on a Neural Network and a Gaussian process regression model, and is used to personalize the estimation of self-reported pain via a set of hand-crafted personal features and multitask learning. 
Xu~\etal~\cite{Xu2020Pain} proposed a three-stage multitask pain model to estimate self-reported pain scores. First, a VGGFace neural network is used to predict frame-level PSPI based pain scores. Second, a fully connected neural network is employed to estimate the VAS at sequence-level from frame-level PSPI predictions using multitask learning to learn multidimensional pain scales instead of the VAS for the entire sequence. Finally, an optimal linear combination of the multidimensional sequence-level VAS was used to predict the final VAS based pain score. 
Xu~\etal~\cite{Xu2020ExploringMM} further refined the work in~\cite{Xu2020Pain} by using the four labels available in the dataset (\ie, VAS, AFF, SEN and OPR) to estimate the level of pain from human-labeled Action Units. The authors combined the use of multitask learning neural network to predict pain scores with an ensemble learning model to linearly combine the multi-dimensional pain scores to estimate the VAS. 
Erekat~\etal~\cite{Erekat2020Enforcing} proposed a spatio-temporal Convolutional Neural Network - Recurrent Neural Network (CNN-RNN) for the automatic measurement of self reported pain and observed pain intensity, respectively. The authors proposed a new loss function that explores the added value of combining different self reported pain scales for a reliable assessment of pain intensity from facial expression. Using an automatic spatio-temporal architecture, they proposed a reliable assessment of pain by maximizing the consistency between different pain assessment scales. Their results show that enforcing the consistency between different self-reported pain intensity scores collected using different pain scales enhances self-reported pain estimation.

All the previously mentioned methods make use of (deep-)neural networks or try to estimate pain intensity at frame level first (PSPI scores), and predict the sequence level pain index from this first estimation. 

Only few works investigated video or geometric based approaches to estimate self-reported pain using the Biovid Heat Pain dataset~\cite{Walter2013Biovid}. In this dataset, the self-reported pain ranges from 0 to 4 (where 0 corresponds to no pain and 4 to high level of pain). This dataset is composed of different parts that come with several modalities such as long or short video sequences and biomedical signals like ECG, EMG or skin conductance. Much of the work that has been done with this dataset used the Part $A$, which comes with biomedical signals and short sequences to estimate the pain intensity at sequence level. Skin conductance was used by Pouromran~\etal~\cite{Pouromran2021BiovidPain} or Lopez-Martinez and Picard~\cite{LopezMartinez2018PainBiovid}. 
Other approaches like the one proposed by Kachele~\etal~\cite{Kachele2017Pain} tested the combination of different modalities. They also extracted the facial landmarks and computed several statistical geometric features from the raw coordinates. In a different way, Lopez-Martinez~\etal~\cite{LopezMartinez2017NIPS} also extracted facial landmarks in order to compute statistical geometric features and combine them with the biomedical signals to estimate the levels of pain. However, they do not use the short video sequences available in the Part $A$ of the dataset.


\begin{figure*}[!ht]
	\centering
	\includegraphics[width=0.9\textwidth]{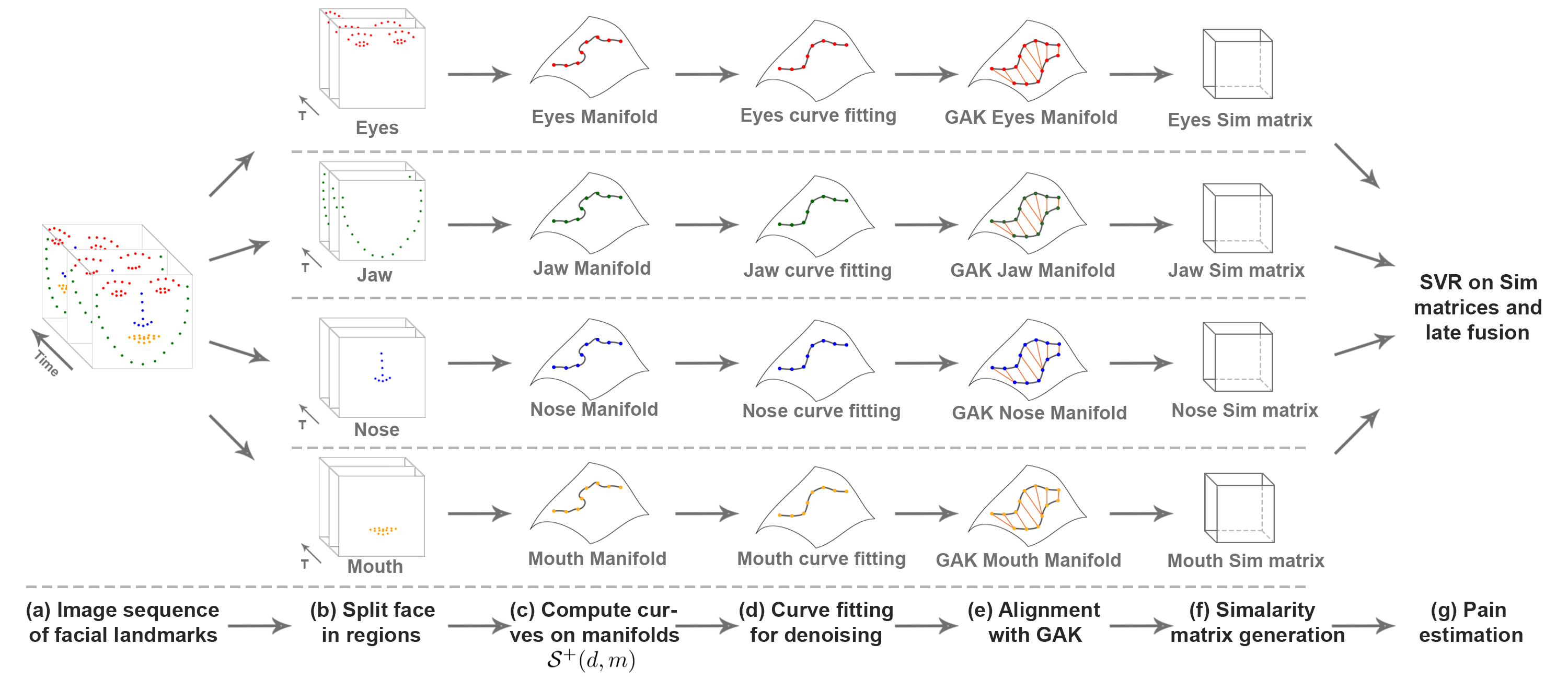}
	\caption{Method overview: \emph{(a)} Detection and extraction of facial landmarks; \emph{(b)} Split of the landmark configurations in different regions and computation of their velocities; \emph{(c)} Computation of Gram matrices and modeling of their temporal dynamics as trajectories on the $\mathcal{S}^+(2,m)$ manifold; \emph{(d)} Application of curve fitting for noise reduction and smoothing of the trajectories; \emph{(e)} Alignment of the trajectories with the Global Alignment Kernel (GAK); \emph{(f)} Similarity matrix computation for all the regions; \emph{(g)} Pain estimation for each region and late fusion of the scores for the final pain level.}
	\label{fig:overview}
\end{figure*}

In this paper, we propose a model for estimating VAS pain intensity score at sequence level by analysing the dynamics of the face. In particular, the face dynamics is described in terms of position and velocity of facial landmarks. To investigate the relevance of different facial regions for pain estimation, facial landmarks are grouped into four clusters corresponding to different anatomical regions of the face.
The dynamics of the landmarks in each region is modeled as a trajectory on the manifold of Positive Semi-Definite Matrices of fixed rank. 
Modeling the temporal evolution of landmarks as a trajectory on a Riemannian manifold has shown promising results in action recognition~\cite{Szczapa2019ICCVW, Devanne2015Cybernetics, amor2016action} and in facial expression recognition~\cite{KacemDABP20}. Motivated by these results, we propose a geometric approach for VAS pain intensity estimation based on the representation of facial landmarks and their dynamics as Gram matrices of fixed rank. The dynamics of the face is formulated as a trajectory on the Riemannian manifold of positive semi-definite matrices of fixed rank (Figure~\ref{fig:overview}).
To cope with noisy data and accommodate for different frame rates of landmark detection, a cubic B\'ezier curve fitting model is adopted to approximate Gram matrix trajectories on the manifold.
Ultimately, distances between trajectories on the manifold are used to build a kernel of similarities, used to train a Support Vector Regression model that estimates the VAS pain index score. Evaluation using two benchmark datasets demonstrates that the proposed solution improves the accuracy compared to the state-of-the-art.
This work develops on the model preliminary proposed in Szczapa~\etal~\cite{szczapa2020automatic}, exploiting the idea of modeling the dynamics of the face with Gram matrices on the manifold of positive semi-definite matrices of fixed rank. Compared to this preliminary approach, the main novelties of this paper are:
\begin{itemize}
	\item A region-based representation of the face is used rather than a holistic one. This enables us to investigate the relevance of different facial parts for pain estimation;
	\item An investigation on how to perform the fusion of the regions to estimate the pain at a sequence level;
	\item The proposed approach is tested on two datasets, showing the effectiveness of facial decomposition.
\end{itemize}

\section{Facial Shape Representation}
\label{sect:face-representation}
We propose a video based measurement of self-reported VAS based pain intensity scores using the dynamics of facial movement. An overview of the proposed approach is reported in Figure~\ref{fig:overview}. First, facial landmarks are detected from each video frame to form a sequence of landmark configurations. The landmark configurations are then split into four regions to form four sequences of facial region landmark configurations. For each region based time series, velocities are then computed as the displacement of the coordinates between two consecutive frames. Gram matrices are computed from the combination of the landmark coordinates and their velocities. These matrices are represented as trajectories on the $\mathcal{S}^+(2,m)$ manifold, which is the set of $m \times m$ symmetric positive semi-definite matrices of rank~2, with one manifold per region. We apply a curve fitting algorithm to the trajectories of each manifold for smoothing and noise reduction. Alignment of the trajectories is obtained by using the Global Alignment Kernel (GAK)~\cite{CuturiVBM07}, which results in a similarity matrix per region containing the similarities between trajectories of homologous regions. Finally, we use the kernels generated by GAK with SVR to estimate the pain intensity based on each region. A late fusion is then used to combine the estimated pain scores. 


\subsection{Gram Formulation} \label{sect:landmarks}

	Given an image sequence, we represent the dynamics of facial movement with a time series formed by the coordinates $\left(x_{1}, y_{1}\right),\left(x_{2}, y_{2}\right), \ldots,\left(x_{n}, y_{n}\right)$ of $n$ tracked facial landmarks and grouped into matrices $Z_i$. Each $Z_i$ ($0 \leq i \leq \tau$) being a $n \times 2$ matrix $\left[\left(x_{1}, y_{1}\right),\left(x_{2}, y_{2}\right), \ldots,\left(x_{n}, y_{n}\right)\right]^{T}$ of rank~2 encoding the positions of the $n$ facial landmarks. 
	For each landmark $l_{i}$, its velocity is also measured as the magnitude of the displacement between two consecutive matrices $Z_{i}$ and $Z_{i+1}$. 
	We denote the velocity matrix at frame $i$ as $V_{i} = Z_{i+1} - Z_{i} \in \R^{n \times 2}$. Since velocity cannot be extracted from the last frame, $V_i$ is computed only for $i\in\{0,\ldots,\tau-1\}$. 
	However, to simplify the notation, we adopt the same range of the frame indexes $\{1,\tau\}$ for both the landmark position and their velocity. In doing so, the last frame is dropped from the video sequence, and it is only used to estimate the velocity. 
	
	%
	
\begin{figure}[!ht]
		\centering
		\includegraphics[width=0.5\linewidth]{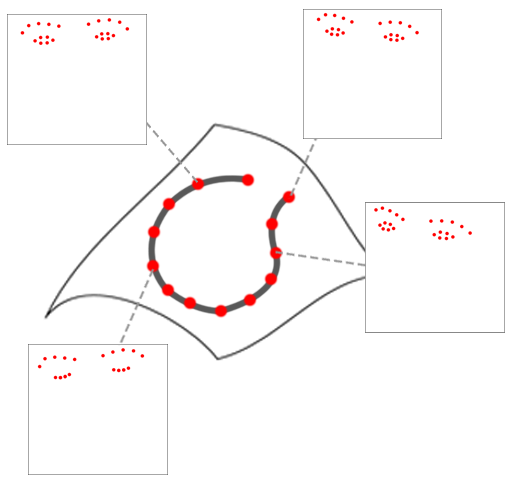}
		\caption{Example of a trajectory of Gram matrices for the eyes region.}
		\label{fig:traj-eyes}
	\end{figure}

	Our objective here is to find a shape representation that is invariant to Euclidean transformations (rotations and translations). To remove the translation, each landmark configuration $Z_{i}$ is centered by subtracting the landmarks center of mass. The velocity of each landmark is computed after this normalization.
	Similar to~\cite{Daoudi19ACII, Szczapa2019ICCVW, KacemDABP20}, we propose the Gram matrix $G$ as a representation of landmarks and velocities. The Gram matrix is defined by:
	\begin{equation}
		\label{eq:gram}
		G = FF^{T} = \left\langle p_{i}, p_{j}\right\rangle, \quad 1 \leq i, j \leq 2n,
	\end{equation}
	
	\noindent
	where $F=[Z|V]$ is the $2n \times 2$ matrix obtained by concatenating the position $Z$, and the velocity $V$ of the landmarks. 
	The Gram matrix representation is invariant to rotation and translation. In addition, Gram matrices of the form $FF^T$, where $F$ is an $m \times 2$ matrix of rank~2 ($m=2n$), are characterized as $m \times m$ positive semi-definite (PSD) matrices of rank~2, a Riemannian manifold of well-studied geometry and theoretical properties~\cite{Bonnabel2009SIAM}. As an example, Figure~\ref{fig:traj-eyes} shows a Gram matrix representation as a trajectory on the manifold of PSD matrices.
	
	\subsection{Gram Matrix Distance}\label{sect:psd}
	To model the dynamic changes of landmarks as the distance between consecutive Gram matrices, we consider here the Riemannian geometry of the space $\psd{2}{m}$ of $m \times m$ positive semi-definite matrices of rank~$2$~\cite{Bonnabel2009SIAM}. This Riemannian geometry has been studied in~\cite{bonnabel:2009, journee2010low, massart2020quotient, massart2019curvature, vandereycken2009embedded, vandereycken2013riemannian} and used in~\cite{faraki:2016, meyer2011regression, gousenbourger2017piecewise, massart2019interpolation}. 
	In~\cite{Szczapa2019ICCVW}, it was demonstrated that the distance between two Gram matrices $G_{i}=F_iF_i^T$ and $G_{j}=F_jF_j^T$ can be defined as follows:
	\begin{equation}
		\label{eq:PSD-norm}
		d(G_{i}, G_{j}) = \mathrm{tr}(G_{i}) + \mathrm{tr}(G_{j}) - 2 \mathrm{tr}\left( \left( G_{i}^{\frac{1}{2}}  G_{j} G_{i}^{\frac{1}{2}} \right)^{\frac{1}{2}} \right).
	\end{equation}
	
	\noindent
	This distance can be expressed in terms of the facial configurations $F_{i}, F_{j} \in \frmat{m}{2}$ as follows:
	\begin{equation}
		\label{eq:frob-norm-new-metric}
		d(G_{i}, G_{j}) = \min_{Q \in \calO_2} \lVert F_{j}Q-F_{i}\rVert_F ,
	\end{equation}
	
	\noindent
	where $\lVert .\rVert_F$ is the Frobenius norm. The optimal solution is $Q^* := VU^{T}$, where $F_{i}^{T} F_{j} = U \Sigma V^{T}$ is a singular value decomposition. 
	In case the facial landmarks are points in a 2D space, \eqref{eq:PSD-norm} can be evaluated in a computationally convenient form, as stated in the following Theorem.
	
	\begin{theorem}
		\label{theorem:distance}
		Considering $G_{i}, G_{j} \in \psd{2}{m}$ as the two Gram matrices obtained from facial configurations $F_{i}, F_{j} \in \mathbb{R}^{m \times 2}$, the Riemannian distance in~\eqref{eq:PSD-norm} can be expressed as:
		\begin{equation}
			\label{eq:distp2}
			d(G_{i}, G_{j})= \Tr(G_{i}) + \Tr(G_{j}) - 2\sqrt{(a + d)^2 + (c - b)^2} \; ,
		\end{equation}
	\end{theorem}
	
	\noindent
	where $F_{i}^{T}F_{j}$ = $\left( \begin{array}{cc} a & b \\ c & d \end{array} \right)$. 
	
	\begin{proof}
	We can reformulate our metric introduced in~\eqref{eq:frob-norm-new-metric} with:
	\begin{align*}
		d^2(G_i, G_j) & = \Tr \left[ (F_jQ - F_i)(F_jQ - F_i)^T \right] \\
		& = \Tr(G_i) - 2\Tr(F_iQ^TF_j^T) + \Tr(G_j) .
	\end{align*}
	
	To minimize our distance, we need to maximize the term $\Tr(F_iQ^TF_j^T)$. Let $F_j^TF_i$ be a $2 \times 2$ matrix with four unknown values $a, b, c, d$ and let $Q \in \mathcal{O}_p$, we  maximize:
	\begin{equation}
		\label{eq:maximize-term}
		\begin{split}
				\max\;\Tr \left[ \left( \begin{array}{cc} a\cos{\Theta} - b\sin{\Theta} & - \\ - & c\sin{\Theta} + d\cos{\Theta} \end{array} \right) \right] .
			\end{split}
	\end{equation}
	
	From~\eqref{eq:maximize-term}, we now have to find the maximum of $(a + d)\cos{\Theta} + (c - b)\sin{\Theta}$, meaning that we have to maximize $\sqrt{(a + d)^2 + (c - b)^2}\cos{(O-O')}$. As we want to maximize this value, $O$ has to be equal to $O'$, so $\sqrt{(a + d)^2 + (c - b)^2}\cos{(O-O')} \leqslant \sqrt{(a + d)^2 + (c - b)^2}$. Therefore we can say that:
	\begin{equation}
		\max \Tr(F_iQ^TF_j^T) = \sqrt{(a + d)^2 + (c - b)^2} .
	\end{equation}
	\end{proof}
	

	\section{Modeling the Temporal Dynamics of Landmarks}\label{sect:face-dynamics}
	Based on the landmark representation introduced in the previous section, each face in a frame of a video sequence is mapped to a point on the PSD manifold. Thus, it becomes natural to interpret the points mapped from consecutive frames as describing a trajectory on the manifold. However, making these trajectories useful for subsequent processing and comparison requires smoothing (Section~\ref{sect:bezier}) and alignment (Section~\ref{sect:global-alignment}), as illustrated in the following.
	
	\subsection{Trajectory Modeling and Smoothing}\label{sect:bezier}
	The dynamic changes of facial landmarks movement originate trajectories on the Riemannian manifold of positive-semidefinite matrices of fixed rank. 
	We fit a curve $\beta_G$ to a sequence of landmark configurations $\{F_0, \ldots, F_\tau\}$ represented by their corresponding Gram matrices $\{G_0,\ldots,G_\tau\}$ in $\psd{2}{m}$. This curve enables us to model the spatio-temporal evolution of the elements on $\psd{2}{m}$. 
	
	Modeling a sequence of landmarks as a piecewise-geodesic curve on $\psd{2}{m}$ showed very promising results when the data are well acquired, \ie, without tracking errors or missing data. 
	To smooth the data, accounting both for missing data and tracking errors, we propose to use cubic blended curve fitting algorithms~\cite{GousenbourgerMM17, GousenbourgerJMIV19}. These algorithms only require to compute Riemannian exponential and logarithm, and also represent the curve by means of a number of tangent vectors that grows linearly with the number of data points. In this paper, we use the algorithm defined in~\cite{Gousenbourger2018}.
	Specifically, given a set of points $\{G_0, \ldots, G_\tau\} \in \psd{2}{m}$ associated to times $\{t_0, \dots, t_\tau\}$, with $t_i := i$, the curve $\beta_G$, defined on the interval $[0,\tau]$, is defined as:
	\begin{equation}
		\beta_G(t) := \gamma_i(t-i), \qquad t \in [i, i+1] ,
	\end{equation}
	
	\noindent
	where each curve $\gamma_i$ is obtained by blending together fitted cubic B\'ezier curves computed on the tangent spaces of the data points $d_i$ and $d_{i+1}$ (represented by Gram matrices on the manifold forming a trajectory). The De Casteljau algorithm, used during the reconstruction process is fully performed in the tangent spaces of $d_i$ and $d_{i+1}$, and a weighted mean is done on the two obtained points.
	
	These fitting cubic B\'ezier curves depend on a parameter $\lambda$, allowing us to balance two objectives: \emph{(i)} Proximity to the data points at the associated time instants; \emph{(ii)} Regularity of the curve (measured in terms of mean square acceleration). A high value of $\lambda$ results in a curve with possibly high acceleration that almost interpolates the data, while taking $\lambda \rightarrow 0$ results in a smooth function approximating the original trajectory.
	
	\subsection{Global Alignment}\label{sect:global-alignment}
	As introduced in the previous section, we represent a sequence of Gram matrices as a trajectory in $\psd{2}{m}$. Since videos could have different duration (\ie, video sequences of pain, in our case), the length of the corresponding trajectories represented in this manifold can be different. 
	The Dynamic Time Warping (DTW) algorithm is a commonly used method to compute the similarity between trajectories of different length. However, DTW does not define a proper metric and cannot be used to derive a valid positive-definite kernel. This would hamper the use of many approaches (including Support Vector Regression) to learn the mapping between trajectories in $\psd{2}{m}$ and pain intensity.
	Cuturi~\etal~\cite{CuturiVBM07} proposed the Global Alignment Kernel (GAK) to address non-positive definite kernels induced by DTW. 
	GAK allows us to derive a valid positive-definite kernel when aligning two time series. As opposed to the DTW, the  kernel generated by GAK, that is the similarity matrix between all the sequences, can be used directly with Support Vector Regression. In fact, the kernels built with DTW do not show favorable positive definiteness properties as they rely on the computation of an optimum, rather than the construction of a feature map. In terms of complexity, similar to naive implementation of DTW, the computational complexity of the GAK kernel is quadratic.
	
	Let us now consider $G^{1} = \{G^{1}_{0}, \cdots, G^{1}_{\tau_1}\}$ and $G^{2}=\{G^{2}_{0}, \cdots, G^{2}_{\tau_2}\}$, two sequences of Gram matrices. Given a metric $d$ to compute the distance between two elements of two sequences (see~\eqref{eq:distp2}), we compute the matrix $D$ of size $\tau_{1} \times \tau_{2}$, where each $D(i,j)$ is the distance between the $i$-th and $j$-th elements of the two sequences, respectively, with $1 \leq i \leq \tau_{1}$ and $1 \leq j \leq \tau_{2}$:
	\begin{equation}
		\label{eq:matrix-distance}
		D(i,j) = d(G^{1}_{i}, G^{2}_{j}) .
	\end{equation}
	
	The kernel $\tilde{k}$ can now be computed using the halved Gaussian Kernel on this same matrix $D$. Therefore, the kernel $\tilde{k}$ can be defined as:
	\begin{equation}
		\label{eq:k-tilde}
		\tilde{k}(i,j) = \frac{1}{2} * exp\left(-\frac{D(i,j)}{\sigma^2}\right) ,
	\end{equation}
	
	\noindent 
	with $0 \le \sigma \leq 1$. 
	The choice of the right value of $\sigma$ can affect the similarity score. If this value is too small, it penalizes the values of the distance matrix $D$, and the similarity scores obtained at the end of the process will be too small and similar to each other, even when two sequences are very different. In the same way, if this value is too high, the values in the kernel $\tilde{k}$ will be similar, and the similarity scores will be too high and also similar to differentiate the sequences. The choice of the best value for the $\sigma$ parameter is explained in Section~\ref{sec:best-configuration}.
	
	As reported in~\cite{CuturiVBM07}, we can redefine our kernel as:
	\begin{equation}
		\label{eq:kernel-PSD}
		k(i,j) = \frac{\tilde{k}(i,j)}{(1-\tilde{k}(i,j))} .
	\end{equation}
	
	\noindent
	This strategy guarantees us the kernel is positive semi-definite and can be used in its own. Finally, we can compute the similarity score between the two trajectories $G^{1}$ and $G^{2}$. This computation is performed with quadratic complexity, like DTW. To do so, we define a new matrix $M$ that contains the path to the similarity between our two sequences. We define $M$ as a zeros matrix of size $(\tau_1+1) \times (\tau_2+1)$ and $M_{0,0} = 1$. Computing the terms of $M$ is done using Theorem~2 in~\cite[\S 2.3]{CuturiVBM07}:
	\begin{equation}
		\label{eq:matrix-similarity}
		M_{i,j} = (M_{i,j-1} + M_{i-1,j-1} + M_{i-1,j})*k(i,j) .
	\end{equation}
	
	\noindent
	The similarity score between the trajectories $G^{1}$ and $G^{2}$ is given by the value at $M_{\tau_1+1,\tau_2+1}^{p}$, being $p\in\{0,1,2,3\}$ the index used to denote one of the four face regions.

	\section{Pain Estimation}\label{sect:pain-estimation}
	Having a representation of landmark sequences as a trajectory on the PSD manifold and a similarity measure between them, we are in the position of using these similarities to train a regressor for VAS pain score estimation at video level.
	
	As described in Section~\ref{sect:face-representation}, the dynamics of the facial landmarks is captured by four trajectories, each one capturing the dynamics of one out of four regions of the face.
	In order to estimate the pain score, these trajectories can be processed following three different strategies: \emph{(i)} perform a manifold product among the four manifolds, one for each region, to form a new valid manifold and compute the similarity between the trajectories on this new manifold; \emph{(ii)} compute the similarity scores between the trajectories on each manifold independently, then perform an early  fusion to estimate the pain score; \emph{(iii)} compute the similarity scores between the trajectories on each manifold independently, then perform a late fusion to estimate the pain score.
	
	Given a dataset composed of $n_{seq}$ videos annotated with pain intensity score, a sequence of facial landmark configuration matrices is extracted from each video. 
	Then, a symmetric matrix $K^{p}$ of size $n_{seq} \times n_{seq}$ is built to store the similarity scores between all the trajectories for a given region $p$. These matrices are built with values computed using the positive semi-definite kernel, meaning that each matrix is positive semi-definite. Now that we have a valid and positive semi-definite kernel $K^{p}$, as demonstrated by Cuturi~\etal~\cite{CuturiVBM07}, we can use it directly as a valid kernel for estimation.
	To estimate the pain intensity score (\ie, self-reported VAS scores), we use a Support Vector Regression (SVR) model~\cite{Smola2004SVM, Drucker1996SMV}. 
	In order to predict the level of pain based on similarity matrices from different regions of the face, three different strategies can be adopted, namely, \textit{manifold product}, \textit{early fusion} and \textit{late fusion}. These three strategies are combined with two evaluation protocols to estimate the pain index of each sequence, \ie, \textit{Leave-One-Subject-Out cross validation} and \textit{$k$-fold cross validation}.
	
	\subsection{Evaluation Protocols}\label{sec:protocols}
	To evaluate the proposed method, we used the two subject-independent protocols presented in~\cite{szczapa2020automatic}: Leave-One-Subject-Out cross validation, and $k$-fold cross validation.
	
	\textbf{Leave-One-Subject-Out cross validation (LOSO)} -- In this protocol, for each round, we use all the sequences from all subjects but two for training, and the remaining sequences of one subject for validation and the sequences of the other subject for testing. There is no overlap between the training, validation and testing sets. Accordingly, this is a \emph{subject-independent} evaluation protocol. We perform this operation for all the subjects in the dataset, so that each subject is used once for testing.
	
	\textbf{$k$-fold cross validation} -- This protocol is similar to the LOSO cross validation one, but instead of taking only the sequences of one subject at a time for validation and testing at each round, we take all the sequences of $k$ subjects for validation and testing, and the remaining sequences for training. The choice of the $k$ subjects for the validation set is done by choosing the $k$ first subjects in the dataset, then the $k$ next subjects for testing and the remaining for training and so on until all the subjects are used for testing (\ie, $k$ rounds). Also this evaluation protocol is subject-independent.
	
	\smallskip
	
	Cross validation has the advantage of preventing from having results that are due to the chance as all data are used to train and test the proposed method. The average across all folds is more representative of the whole dataset.
	
	\subsection{Regions Manifold Product}
	The idea here to compute pain scores is that of combining the manifolds, one for each region, before using SVR for pain estimation. Indeed, the decomposition of the face into four regions can be seen as the product space of four manifolds $\mathcal{M} = \psd{2}{n_1} \times \psd{2}{n_2} \times \psd{2}{n_3} \times \psd{2}{n_4}$, one manifold per region. Thus, the distance between two elements $G_{i}, G_{j} \in \mathcal{M}$ can be modeled as the square root of the sum of the squared distances between these elements in each manifold~\cite{Boothby:107707}:
	\begin{equation}
		\label{eq:distprod}
		d_{\mathcal{M}}(G_{i}, G_{j}) = \sqrt{ \sum_{k=1}^{4}d(G_{ki}, G_{kj})^{2}_{\psd{2}{n_k}}} \; ,
\end{equation}

\noindent
where each $n_k$, $1 \le k \le 4$, encodes the landmark coordinates in region $k$, and $d(., .)$ is the distance defined in~\eqref{eq:distp2}.
The result is a new manifold that preserves the structure of the original manifolds. 
The result of this formula is the distance between $G_{i}$ and $G_{j}$. This distance is computed between all the Gram matrices composing two trajectories and the alignment algorithm is then performed on the resulting distance matrix as explained in the previous section.

The distances between the trajectories in the manifold $\mathcal{M}$ are computed to form the similarity matrix. As the manifolds are combined into one new manifold, only one similarity matrix is computed and used as our kernel for estimation. In this case, no weights combination is performed and we only have to train one SVR, like in the early fusion strategy.

\subsection{Early Fusion}
In this strategy, the SVR model is fed with the combination of the four kernels $K^p$ and trained to estimate pain score. By adopting the early fusion approach, the combination of the kernels is done by averaging the similarity scores:
\begin{equation}
	\label{eq:early-fusion}
	K_{i,j} = \frac{\sum_{p=1}^{4} K^{p}_{i,j}}{4} .
\end{equation}

\noindent
By doing so, we only need to train one SVR for the whole face using this new kernel that is computed by fusing the scores of different regions of the face in such a way that all regions are assigned the same weight.

\subsection{Late Fusion}\label{sec:late-fusion}
When adopting the late fusion strategy, the training sets used as inputs to train the models are part of our kernel $K^{p}$ containing the similarity scores between all the training trajectories for the region $p$. Taking a subset of the entire kernel $K^{p}$ for training gives us a new kernel that is also positive semi-definite by construction. A vector containing the ground-truth VAS scores for the trajectories is also given for the training part. Finally, the outputs of region specific models are combined to predict the VAS scores for the whole face.
Accordingly, we train one SVR per region independently, using the kernels $K^{p}$. Once the VAS scores are predicted for all the regions and for all the sequences in the dataset, we apply a late fusion of the scores to obtain the VAS pain index $\mathrm{\hat{y}}$ for the whole face by taking a weighted combination of the four predictions for each sequence: 
\begin{equation}
	\label{eq:late-fusion}
	\mathrm{\hat{y}} = \frac{\left(w_j \cdot \hat{y}_{jaw} + w_n \cdot \hat{y}_{nose} + w_m \cdot \hat{y}_{mouth} + w_e \cdot \hat{y}_{eyes}\right)}{4} . 
\end{equation}

\noindent
In order to identify the best combination of the weight values, a grid search approach has been adopted, with values in the set $\left\{0.1, 0.2, \ldots, 0.9, 1.0 \right\}$. The best weight values are determined at each round of the cross validation by taking out the sequences of one (or $k$) subject that will be used as testing data. Then, a second cross validation loop is included inside the first one, where the sequences of a subject are taken out and used as validation data, while the remaining sequences are used as training data. The weights are estimated at each round of this second cross validation loop, using the validation data, and the best weight combination is used to estimate the pain index of each sequence of the testing data. By this double cross validation loop, the weights are optimized using validation data that are not included in the testing set, reducing the risk of overfitting.

\section{Experimental Results}\label{sect:results}
Our goal here is to estimate self reported pain intensity from videos. To do so, the proposed approach has been experimented on two benchmark datasets for pain detection: the UNBC-McMaster Shoulder Pain Archive~\cite{UNBC-McMaster} and the Biovid Heat Pain dataset~\cite{Walter2013Biovid}.
Description of the data, the adopted protocols, and the results are reported in Section~\ref{sect:UNBC-McMaster-dataset} and Section~\ref{sect:Biovid-dataset}, respectively, for the two datasets.

Since the prediction of the VAS score is a continuous value, the evaluation of our approach is obtained by computing two error measures: the Mean Absolute Error (MAE) and the Root Mean Square Error (RMSE) between the predicted pain scores and the ground-truth. 
The MAE is computed as follows:
\begin{equation}
	\mathrm{MAE} = \frac{1}{n_{seq}} \sum_{i=1}^{n_{seq}}\left|y_{i} - \hat{y_{i}}\right| ,
\end{equation}

\noindent
and the RMSE is given by:
\begin{equation}
	\mathrm{RMSE}=\sqrt{ \frac{\sum_{i=1}^{n_{seq}}\left( y_{i} - \hat{y_{i}} \right)^{2}} {n_{seq} }} ,
\end{equation}

\noindent
where $n_{seq}$ is the number of sequences considered, $y_{i}$ is the ground truth (\ie, self-reported VAS pain score), and $\hat{y_{i}}$ is the predicted pain score.

\subsection{The UNBC-McMaster Shoulder Pain Archive}\label{sect:UNBC-McMaster-dataset}
\begin{figure}[!ht]
	\centering
	\subfloat[]{\label{fig:frame-34}\includegraphics[width=0.35\linewidth]{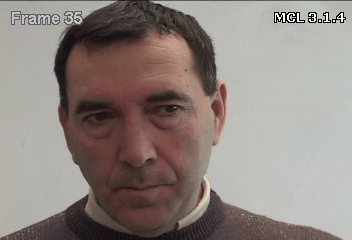}}
	~
	\subfloat[]{\label{fig:frame-34-landmarks}\includegraphics[width=0.38\linewidth]{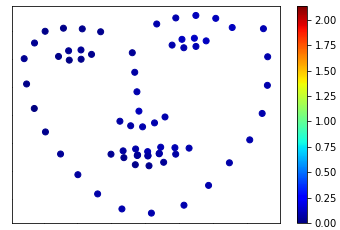}}
	~ \\
	\subfloat[]{\label{fig:frame-289}\includegraphics[width=0.35\linewidth]{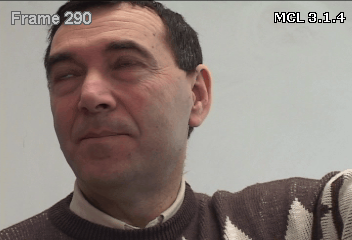}}
	~
	\subfloat[]{\label{fig:frame-289-landmarks}\includegraphics[width=0.38\linewidth]{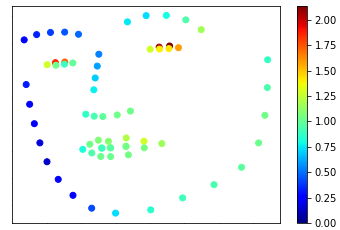}}
	\caption{UNBC-McMaster Shoulder Pain Archive~\cite{UNBC-McMaster}: (a) and (c) show two example images from a sequence; In (b) and (d) the landmark coordinates for images in (a) and (b) are reported, with velocities evidenced by different colors (best viewed in color).}
	\label{fig:example-frames-dataset}
\end{figure}

The UNBC-McMaster Shoulder Pain Archive~\cite{UNBC-McMaster} is a widely used dataset for pain expression recognition and intensity estimation. It contains 200 video recordings of 25 subjects performing active and passive range-of-motion of their affected and unaffected shoulders. Each video sequence is annotated for pain intensity score at the sequence-level using three self-reported scales (including the VAS) and an Observer Pain Rating scale. The video recordings are also annotated at the frame-level using the manual FACS. The facial landmarks are available with the dataset and are extracted using an Active Appearance Model (AAM). In total, 66 landmarks are available at the jaw, the mouth, the nose, the eyes and the eyebrows. Figure~\ref{fig:example-frames-dataset} shows two images from a sequence of the dataset with their corresponding facial landmarks, colored by their velocities. Our goal is to estimate the self-reported pain score (VAS). 

\begin{figure}[!ht]
	\centering
	\includegraphics[width=0.55\linewidth]{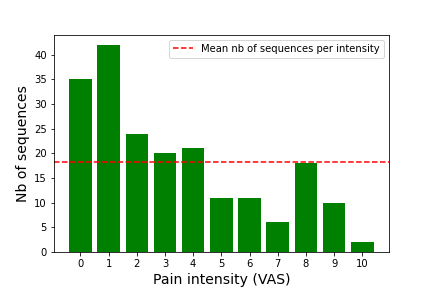}
	\caption{Distribution of the VAS Pain Scores for the UNBC-McMaster Shoulder Pain Archive.}
	\label{fig:videos-per-intensity}
\end{figure}

\begin{figure}[!ht]
	\centering
	\includegraphics[width=0.55\linewidth]{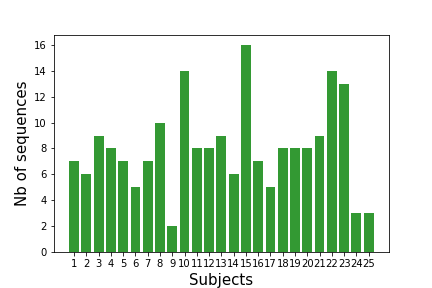}
	\caption{Number of sequences per subject for the UNBC-McMaster Shoulder Pain Archive}
	\label{fig:videos-per-subjects}
\end{figure}

Figure~\ref{fig:videos-per-intensity} shows the distribution of the VAS score across the dataset. One can observe that the number of available sequences per VAS score is not uniformly distributed: $50$\% of the sequences have a VAS pain score of $\{0,1,2\}$, while only $11$\% of the sequences have a VAS pain score of $\{8,9,10\}$. 
Also, the number of sequences per subject is not uniform, as shown in Figure~\ref{fig:videos-per-subjects}. This bias, both in terms of number of sequences per VAS score and number of sequences per subject, hampers accurate learning and prediction of the VAS score, making the estimation more challenging. 

\subsubsection{Ablation Study and Best Configuration Estimation}\label{sec:best-configuration}

The process of estimating the pain index by the analysis of face dynamics depends on three main hyper-parameters that determine the amount of smoothing of trajectories on the manifold (\ie lambda value), the sigma value used as a parameter for the application of the Gaussian Kernel during sequence alignment, and the number of frames that are actually used to compute these trajectories. 
In fact, reducing the number of frames for each sequence allows us to speed up the computation time because we need to compare fewer frames to calculate the similarity score between any two sequences in the dataset. 
To identify a convenient choice of these hyper-parameters, a grid search strategy is adopted. For this purpose, the value of the parameter lambda (Section~\ref{sect:bezier}) is discretized into four reference values $\{No fitting, 10, 100, 1000\}$, with $10$ meaning we apply a fairly strong amount of smoothing of the trajectories and $1000$ a soft application of smoothing (this value is closer to no fitting than $10$) and $100$ as a middle value for smoothing. 
The $\sigma$ parameter is discretized into three reference values $\{0.5, 0.7, 0.9\}$. Explanations on the choice of this parameter can be found in Section~\ref{sect:global-alignment}.
As for the number of frames that are used to compute the trajectory, three different frame subsampling rates were explored: $25\%$, $50\%$ and $100\%$ of the frames, with $25\%$ meaning that we kept only 1 frame out of 4 and $50\%$ meaning that we kept 1 frame out of 2.

The best configuration was identified by computing the prediction accuracy on the validation set of the UNBC-McMaster Shoulder Pain Archive using the LOSO and a 5-fold cross validation protocol as described in Section~\ref{sec:protocols}. The late fusion method was used to estimate the pain scores, as presented in Section~\ref{sec:late-fusion}. Table~\ref{tab:validation-configuration-results-unbc-loso} reports the prediction accuracy in terms of MAE for the different configurations using the LOSO protocol, and Table~\ref{tab:validation-configuration-results-unbc-5fold} reports the accuracy using the 5-fold cross validation protocol.

\begin{table}[!ht]
	\centering
	\caption{MAE of our proposed method on a validation set on the UNBC-McMaster Shoulder Pain Archive using the LOSO protocol. Best results for a given configuration at varying sampling rates is given in bold. The best result is marked with *.}
	\label{tab:validation-configuration-results-unbc-loso}
	\small
	\begin{tabular}{l|c|c|c|c}
		\multicolumn{2}{c}{}              & \multicolumn{3}{c}{Sampling} \\ \hline
		$\sigma$                & $\lambda$     & 25\%    & 50\%    & 100\%    \\
		\hline \hline
		\multirow{4}{*}{0.5} & No fitting & 1.82    & 1.79    & \textbf{1.72}     \\ \cline{2-5}
		& 10         & 1.89    & \textbf{1.72}    & 1.76     \\ \cline{2-5}
		& 100        & 1.75    & \textbf{1.72}    & \textbf{1.72}     \\ \cline{2-5}
		& 1000       & 1.93    & 1.81    & \textbf{1.74}     \\ \hline
		\multirow{4}{*}{0.7} & No fitting & 1.72    & \textbf{1.69}    & 1.74     \\ \cline{2-5}
		& 10         & 1.68    & \textbf{1.66}    & 1.72     \\ \cline{2-5}
		& 100        & 1.65    & \textbf{1.63*}    & 1.66     \\ \cline{2-5}
		& 1000       & 1.71    & 1.67    & \textbf{1.66}     \\ \hline
		\multirow{4}{*}{0.9} & No fitting & 1.74    & \textbf{1.72}    & 1.73     \\ \cline{2-5}
		& 10         & 1.81    & 1.72    & \textbf{1.71}     \\ \cline{2-5}
		& 100        & 1.74    & 1.73    & \textbf{1.70}     \\ \cline{2-5}
		& 1000       & \textbf{1.71}    & 1.79    & 1.88     \\
		\Xhline{2\arrayrulewidth}
	\end{tabular}
\end{table}

\begin{table}[!ht]
	\centering
	\caption{MAE of our proposed method on a validation set on the UNBC-McMaster Shoulder Pain Archive using the 5-fold cross validation protocol. Best results for a given configuration at varying sampling rates is given in bold. The best result is marked with *.}
	\label{tab:validation-configuration-results-unbc-5fold}
	\small
	\begin{tabular}{l|c|c|c|c}
		\multicolumn{2}{c}{}              & \multicolumn{3}{c}{Sampling} \\ \hline
		$\sigma$                & $\lambda$     & 25\%    & 50\%    & 100\%    \\
		\hline \hline
		\multirow{4}{*}{0.5} & No fitting & 1.78    & \textbf{1.69}    & 1.76     \\ \cline{2-5}
		& 10         & 1.70    & \textbf{1.68}    & 1.76     \\ \cline{2-5}
		& 100        & 1.78    & \textbf{1.75}    & \textbf{1.75}     \\ \cline{2-5}
		& 1000       & 1.75    & \textbf{1.73}    & 1.74     \\ \hline
		\multirow{4}{*}{0.7} & No fitting & \textbf{1.65}    & 1.69    & 1.75     \\ \cline{2-5}
		& 10         & 1.76    & \textbf{1.71}    & 1.77     \\ \cline{2-5}
		& 100        & 1.77    & \textbf{1.72}    & 1.76     \\ \cline{2-5}
		& 1000       & 1.75    & \textbf{1.68*}    & 1.79     \\ \hline
		\multirow{4}{*}{0.9} & No fitting & 1.82    & 1.74    & \textbf{1.72}     \\ \cline{2-5}
		& 10         & \textbf{1.82}    & 1.89    & 1.89     \\ \cline{2-5}
		& 100        & 1.86    & \textbf{1.77}    & 1.85     \\ \cline{2-5}
		& 1000       & 1.86    & \textbf{1.78}    & 1.82     \\
		\Xhline{2\arrayrulewidth}
	\end{tabular}
\end{table}

Results in Table~\ref{tab:validation-configuration-results-unbc-loso} show that the best configuration on the validation set corresponds to $\lambda$=$100$ (\ie, soft smoothing of the trajectories), a $\sigma$=$0.7$ (\ie, trade-off between high and low values that can penalize the similarity scores), with a sub sampling of $50\%$ when using the Leave-One-Subject-Out protocol. 
Results in Table~\ref{tab:validation-configuration-results-unbc-5fold} show the same trend using the 5-fold cross validation protocol. The choice of the $\sigma$ value demonstrates that a value too high (\ie. close to $1$) or too low can negatively impact the estimation of the pain index. For the sampling of the sequences, using the total amount of available frames did not improve the results. This can be explained by the fact that $80\%$ of the frames in this dataset are non pain frames. A reduction of the frame sampling rate is also beneficial to the overall computation time as a lower number of frame comparisons is necessary to estimate the similarity between two sequences.



\subsubsection{Results on the Testing Set}
Our goal here is to estimate the VAS pain score for each video sequence. We tested our method with the two protocols described above: the LOSO, and a 5-fold cross validation. Results are reported in Table~\ref{tab:comparison-results-unbc}. We chose to use 5-folds for the $k$-folds cross validation protocol as reported in the state-of-the-art for better comparison.

For each protocol, we fixed the value of the parameters according to the results reported in Section~\ref{sec:best-configuration}: curve fitting parameter lambda ($\lambda$ in Section~\ref{sect:bezier}) equal to $100$, because the data are well acquired, and we do not need a strong smoothing of the curves and a sub sampling of $50\%$. Columns in Table~\ref{tab:comparison-results-unbc} have the following meaning: \emph{Protocol} indicates the protocol used for training and testing our method; \emph{MAE} and \emph{RMSE} are the two error measures of our estimation. 
Furthermore, we report results for the whole face as baseline for comparison.

\begin{table*}[!ht]
	\centering
	\caption{Prediction accuracy of the proposed method on the test set of the UNBC-McMaster Shoulder Pain Archive. Bold values indicate best results without using augmentation. Underlined values have been obtained using augmentation.} 
    \label{tab:comparison-results-unbc}
    \small
    \begin{adjustbox}{width=0.52\linewidth}
        \begin{tabular}{l|c|c|c}
        	\textbf{Protocol} & \textbf{Regression Setup} & \textbf{MAE} & \textbf{RMSE} \\ 
        	\hline \hline
        	\multirow{4}{*}{LOSO cross validation} & Whole Face & 2.52 & 3.27 \\
        	\cline{2-4} & Cartesian product & 2.12 & 2.72 \\
        	\cline{2-4} & Early fusion & 2.39 & 3.11 \\
        	\cline{2-4} & Late fusion & \textbf{1.61} & \textbf{2.04} \\
        	\cline{2-4} & Late fusion - augmented & \underline{1.41} & \underline{1.87} \\ 
        	\Xhline{2\arrayrulewidth}
        	\multirow{4}{*}{5-fold cross validation} & Whole Face & 2.44 & 3.15 \\
        	\cline{2-4} & Cartesian product & 2.28 & 3.02 \\
        	\cline{2-4} & Early fusion & 2.32 & 3.10 \\
        	\cline{2-4} & Late fusion & \textbf{1.59} & \textbf{1.98} \\
        	\cline{2-4} & Late fusion - augmented & \underline{1.36} & \underline{1.75} \\
        	\Xhline{2\arrayrulewidth}
        \end{tabular}
    \end{adjustbox}
\end{table*}

From Table~\ref{tab:comparison-results-unbc}, we notice the best MAE was obtained with the late fusion strategy and the 5-fold cross validation protocol, with an error of $1.59$. The best MAE with the LOSO protocol is $1.61$, also obtained with the late fusion strategy. 
The weights for the late fusion were estimated during the cross-validation rounds on the validation set as mentioned in Section~\ref{sec:late-fusion}. Values of the weights are as follow: $0.39$ for the jaw region, $0.56$ for the nose region, $0.88$ for the mouth region and $0.94$ for the eyes region. The relative values of these weights can be regarded as an index of how much relevant is each part of the face for the prediction of the pain level. The relevance of the eyes region is 34\%, the mouth region 32\%, the nose region 20\% and the jaw region 14\%. For every tested protocol, we obtained a better MAE using facial decomposition compared to the baseline using the whole face. The late fusion approach gives better results than the early fusion strategy. Therefore, training one SVR for each region is more effective than combining the similarity matrices of the regions in one similarity matrix that represents the whole face and train one SVR on that. This observation is also valid for the Cartesian product of the manifolds, where one SVR is trained after the computation of the similarity matrix between the trajectories in the result of the manifold product. However, the manifold product strategy yields better results than early fusion. We report the RMSE as a second error measure of our estimation. Results show the same trend as for the MAE with the best RMSE observed while applying the late fusion strategy.

\begin{figure}[!ht]
\centering
\includegraphics[width=0.55\linewidth]{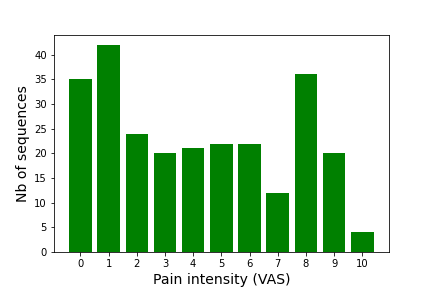}
\caption{VAS pain scores distribution after data augmentation for the UNBC-McMaster Shoulder Pain Archive.}
\label{fig:unbc-augmentation}
\end{figure}

To cope with the non-uniform distribution of videos per class of pain level on the prediction accuracy, we augmented the number of videos of the pain classes where the number of sequences is below the mean number of sequences per class, represented by the red dashed line in Figure~\ref{fig:videos-per-intensity} (the classes concerned are $\{5,6,7,8,9,10\}$). Accordingly, data for sequences with VAS label greater than or equal to 5 were augmented by first flipping the landmark coordinates along the horizontal axis, \ie, $x$ coordinate; Then, each sequence was modeled as a trajectory on the manifold by also applying curve fitting to it. This augmentation allows us to have 58 new sequences with high level of pain (above 5). The new data distribution can be seen in Figure~\ref{fig:unbc-augmentation}. Augmenting the data in this way allowed us to improve the prediction accuracy for all the testing protocols (see underlined scores in Table~\ref{tab:comparison-results-unbc}). We improved our results of about $15$\%, leading to a MAE of $1.36$ for the 5-fold cross validation protocol, and a MAE of $1.41$ with the LOSO protocol. 

\begin{figure}[!ht]
\centering
\includegraphics[width=0.55\linewidth]{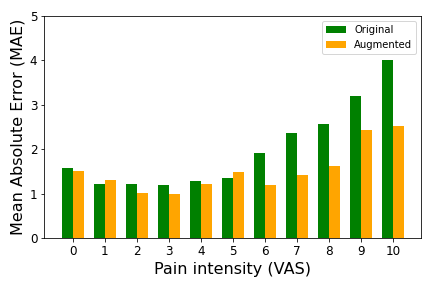}
\caption{MAE per intensity for the LOSO protocol. Green bars represent original data, orange bars represent augmented data.}
\label{fig:mae-per-intensity-unbc-loso}
\end{figure}

\begin{figure}[!ht]
\centering
\includegraphics[width=0.55\linewidth]{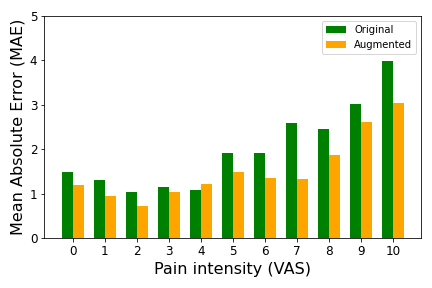}
\caption{MAE per intensity for the 5-fold cross validation protocol. Green bars represent original data, orange bars represent augmented data.}
\label{fig:mae-per-intensity-unbc-5folds}
\end{figure}

Figure~\ref{fig:mae-per-intensity-unbc-loso} and Figure~\ref{fig:mae-per-intensity-unbc-5folds} show the MAE per intensity with the LOSO and 5-folds cross validation protocols, respectively, with and without data augmentation (\ie, green bars show the MAE from original data and orange bars show the MAE from augmented data). From both figures, we can notice that the higher the VAS score is, the higher the MAE is. This can be explained by the fact that there is a limited amount of sequences with high pain scores, as reported in Figure~\ref{fig:videos-per-intensity}. It is also worth noticing that augmenting the data in the dataset, as described above, significantly reduces the MAE per intensity for the sequences with a higher VAS, highlighting the fact that having a more balanced dataset can improve the prediction accuracy.


\paragraph*{\textbf{OpenPose Landmarks:}} We also tested our approach on the UNBC-McMaster dataset using landmarks extracted with OpenPose~\cite{cao2017realtime}. We decided to test two landmark configurations extracted with OpenPose: \emph{(i)} the complete configuration containing $70$ facial landmarks, and \emph{(ii)} a reduced configuration with $66$ facial landmarks corresponding to the original AAM landmarks available with the dataset. We tested these two configurations using the late fusion approach on the two testing protocols. Using the complete OpenPose configuration (\ie, $70$ landmarks), we obtained a MAE of $1.63$ for the LOSO protocol, and a MAE of $1.62$ for the 5-fold cross validation protocol. 
Using the reduced configuration (\ie, $66$ landmarks by excluding the center of the eyes and the corners of the mouth to correspond to the original landmarks available with the dataset), we obtained MAE of $1.62$ and $1.60$ for the LOSO and the 5-fold cross validation protocols, respectively. These results are close to those obtained with the AAM landmarks, indicating that extracted landmarks from a fully automatic method can lead to good results. However, the results are slightly less accurate than those obtained with AAM landmarks. This can be explained by the fact that AAM landmarks are extracted with human in the loop and therefore can be a little more precise than landmarks extracted from a fully automatic method.

\begin{table*}[!ht]
	\centering
	\caption{Comparison of prediction accuracy using different landmarks: the 66 landmarks provided with the UNBC dataset (original), the 70 landmarks provided by the OpenPose library (complete), the 66 landmarks provided by the OpenPose library and corresponding to those provided with UNBC (reduced). Bold values indicate best results.}
    \label{tab:comparison-results-unbc-openpose}
    \small
    \begin{adjustbox}{width=0.5\linewidth}
        \begin{tabular}{l|c|c|c}
        	\textbf{Protocol} & \textbf{Adopted Landmarks} & \textbf{MAE} & \textbf{RMSE} \\ 
        	\hline \hline
        	\multirow{4}{*}{LOSO cross validation} & UNBC Original & \textbf{1.61} & \textbf{2.04} \\
        	\cline{2-4} & OpenPose (complete) & 1.63 & 2.07 \\
        	\cline{2-4} & OpenPose (reduced) & 1.62 & 2.05 \\
        	\Xhline{2\arrayrulewidth}
        	\multirow{4}{*}{5-fold cross validation} & UNBC Original & \textbf{1.59} & \textbf{1.98} \\
        	\cline{2-4} & OpenPose (complete) & 1.62 & 2.00 \\
        	\cline{2-4} & OpenPose (reduced) & 1.60 & 1.98 \\
        	\Xhline{2\arrayrulewidth}
        \end{tabular}
    \end{adjustbox}
\end{table*}

\begin{table*}[!hb]
    \centering
    \caption{Computation time (sec.) for each step of the proposed method on the UNBC-McMaster Shoulder Pain Archive.}
    \label{tab:computation-time-unbc}
    \small
    \begin{adjustbox}{width=0.82\linewidth}
        \begin{tabular}{l|c|c|c|c|c}
        	Sampling & Traj. Comp. (fitting) & Traj. Comp. (no fitting) & Similarity Computation & SVR Training   & Prediction       \\ \hline \hline
        	25\%     & $\approx$ 11.2        & $\approx$ 0.83           & $\approx$ 734          & $\approx$ 0.625 & $\approx$ 0.009 \\ \hline
        	50\%     & $\approx$ 42.7        & $\approx$ 1.56           & $\approx$ 3211         & $\approx$ 0.625 & $\approx$ 0.009 \\ \hline
        	100\%    & $\approx$ 207         & $\approx$ 2.97           & $\approx$ 12422        & $\approx$ 0.625 & $\approx$ 0.009 \\ \Xhline{2\arrayrulewidth}
        \end{tabular}
    \end{adjustbox}
\end{table*}

\paragraph*{\textbf{Computation Time:}} In Table~\ref{tab:computation-time-unbc}, we also summarize the computation time for each step of our approach, with the different sub-samplings and with or without the application of the curve fitting algorithm. Testing is performed on the entire dataset with the LOSO protocol and the late fusion pain estimation, after the estimation of the best combination of weights for each region. Tests were conducted on a laptop equipped with a 6 cores CPU, 16GB RAM, running MatLab 2020b. Table columns have the following meaning: \emph{Sampling} indicates the number of frames that are kept in each sequence of the dataset; \emph{Trajectory Computation} corresponds to the computation of the Gram matrices, trajectory modeling (separate columns are used to report data corresponding to the adoption or not of the curve fitting algorithm); \emph{Similarity Computation} indicates the time to compute the similarity scores between all the sequences in the dataset, including the computation of the distance matrix between all frames of two sequences and the application of GAK; \emph{SVR Training} corresponds to the time to train the four SVR models, one per region, from the similarity matrix, and \emph{Prediction} is the time to predict the self-reported pain score.
From Table~\ref{tab:computation-time-unbc}, we can see the impact of reducing the number of frames for each sequence, especially to build the trajectories on the manifolds and on the computation of the similarity matrix. In fact, each trajectory contains less points as we reduce the number of frames, so a lower number of distance computations is required to measure the similarity between two sequences. We can also note that the application of the curve fitting algorithm can have a strong impact on the computation time. This impact is more significant when we use all the available frames, further demonstrating that processing the video sequences at a reduced frame rate yields computational savings without affecting the prediction accuracy. 
However, applying the fitting algorithm or the sub-sampling of the sequences does not impact the computation time for the SVR training or the prediction of the pain scores. This behavior is desired, as the size of the similarity matrix used for SVR training remains the same (\ie, a square matrix of size $n_{seq} \times n_{seq}$, with $n_{seq}$ the number of sequences in the dataset).

\subsubsection{Comparison with state-of-the-art}
We compared our approach to several state-of-the-art methods for VAS pain intensity measurement from videos (see Table~\ref{tab:comparison-results-sota}). We focused our comparison with other approaches that estimated the pain index at sequence level, but we also reported some results of methods estimating pain index at frame level. The main difference between the two strategies is the use of a different label for training (VAS for sequence level, and PSPI for frame level estimation) and the amount of data used. In order to estimate pain at sequence level, we have to rely on 200 annotated sequences, whereas pain estimation at frame level can leverage on the use of 48,398 annotated frames. 

\begin{table*}[!ht]
\centering
\caption{Comparison of our method with state-of-the-art approaches on the UNBC-McMaster Shoulder Pain Archive. (*~indicates methods that use a neural network)} 
\label{tab:comparison-results-sota} 
\small
\begin{adjustbox}{width=0.82\linewidth}
	\begin{tabular}{l|c|c|c|c|c|c}
		\textbf{Pain Estimation} & \textbf{Method} & \textbf{Protocol} & \textbf{Modalities} & \textbf{Training labels} & \begin{tabular}[c]{@{}l@{}} \textbf{MAE}\\ (VAS)\end{tabular} & \begin{tabular}[c]{@{}l@{}} \textbf{MAE}\\ (PSPI)\end{tabular} \\ 
		\hline \hline
		\multirow{2}{*}{Frame Level} & Deep Pain\cite{Rodriguez2017DeepPain}* & LOSO & Images & PSPI & - & 0.50 \\
		\cline{2-7} & Compact CNN\cite{Semwal2021PainCNN}* & LOSO & Images \& Landmarks & PSPI & - & 0.20 \\
		\Xhline{2\arrayrulewidth}
		\multirow{7}{*}{Sequence Level} & RNN-HCRF~\cite{Martinez:cvrpw17}* & Random split & Facial landmarks & VAS \& PSPI & 2.46 & - \\
		\cline{2-7} & CNN-RNN~\cite{Erekat2020Enforcing}* & 5-fold CV & Images & \begin{tabular}[c]{@{}l@{}}  VAS \& OPI \&\\ AFF \& SEN\end{tabular} & 2.34 & - \\
		\cline{2-7} & DeepFaceLift~\cite{LiuPRP17-DeepFaceLIFT}* & 5-fold CV & Facial landmarks & VAS & 2.30 & - \\
		\cline{2-7} & Extended MTL from pixel~\cite{Xu2020Pain}* & 5-fold CV & Images & VAS & 1.95 & - \\
		\cline{2-7} & Extended MTL with AU~\cite{Xu2020ExploringMM}* & 5-fold CV & Action Units sequences & VAS & 1.73 & - \\
		\cline{2-7} & Manifold trajectories~\cite{szczapa2020automatic} & 5-fold CV & Facial landmarks & VAS & 2.44 & - \\
		\cline{2-7} & \bf{Proposed} & \bf{5-fold CV} & Facial landmarks & \bf{VAS} & \bf{1.59} & - \\
		\Xhline{2\arrayrulewidth}
	\end{tabular}
\end{adjustbox}
\end{table*}

Here, we report the best results for DeepFaceLift~\cite{LiuPRP17-DeepFaceLIFT} for the case where only the VAS scores were used as training labels (in that work authors also presented results, while combining VAS and OPI labels). They obtained a MAE of $2.3$ using a 5-fold cross validation protocol. Our best result for MAE with the same protocol is $1.59$, while only using a geometry based formulation of the dynamics of facial landmarks. 
We also compare our results to the RNN-HCRF method~\cite{Martinez:cvrpw17}. In that work, authors used a different protocol for testing as data have been randomly split by taking the sequences of 15 subjects for training and the remaining 10 sequences for testing. They also used two different labels, the VAS and the PSPI (frame-level label), to train their network to estimate pain at sequence-level. They obtained a MAE of $2.46$ with this configuration. It is important to highlight that in~\cite{Martinez:cvrpw17} the authors used the face appearance, while our method only considers the shape of the face through facial landmarks.
The manifold trajectories proposed in~\cite{szczapa2020automatic} allow the authors to obtain a MAE of $2.44$ when they performed the 5-fold cross validation protocol and a MAE of $2.52$ using the Leave-One-Sequence-Out protocol. Our approach is based on the same structure, but we estimate the self-reported pain level by decomposing the face, whereas in~\cite{szczapa2020automatic} the estimation was performed on the whole face, demonstrating the effectiveness of our proposed facial decomposition. 
Recently, Xu~\etal~\cite{Xu2020Pain} obtained a MAE of $1.95$ using the 5-fold cross validation protocol and this result was further refined in~\cite{Xu2020ExploringMM} with a MAE of $1.73$, using the same protocol. In the first work, the authors estimated the frame-level label before estimating the sequence-level pain. In the second work, they used the different labels available in the UNBC-McMaster Shoulder Pain Archive to estimate the VAS at sequence-level.
Finally, we report the best results for CNN-RNN~\cite{Erekat2020Enforcing}, when the authors combined different labels for training. A MAE of $2.34$ was obtained using a two-level 5-fold cross validation scheme.

\subsection{The Biovid Heatpain Dataset}\label{sect:Biovid-dataset}
The Biovid Heat Pain dataset~\cite{Walter2013Biovid} is widely used for pain expression recognition and pain intensity estimation. This dataset contains 8,700 videos of 87 different subjects. The dataset is composed of 5 pain classes (pain level from 0 to 4), with 20 samples per class and subject, with a time window of 5.5 seconds. The dataset consists of 5 different parts, containing pain stimulation (parts $A$, $B$ and $C$), posed expression (part $D$) and emotion elicitation (part $E$). We worked with part $A$ of the dataset, characterized by the absence of electromyography sensors (EMG) on the user face. In the videos, the subjects are asked to put a hand on a heat source, while the heat sensation increases with the time lapse. The thresholds for the minimum and maximum temperature is determined for each subject on a scale of $\{0,\dots,4\}$, with 0 meaning no pain and 4 meaning worst possible pain level.
Our goal is to estimate the pain intensity scores consistently with the self-reported pain level over the dataset. This dataset is larger and more balanced than the UNBC-McMaster Shoulder Pain Archive, as every subject has 100 sequences, with 20 sequences per pain class.

\begin{figure}[!ht]
\centering
\subfloat[]{\label{fig:biovid-frame}\includegraphics[width=0.35\linewidth]{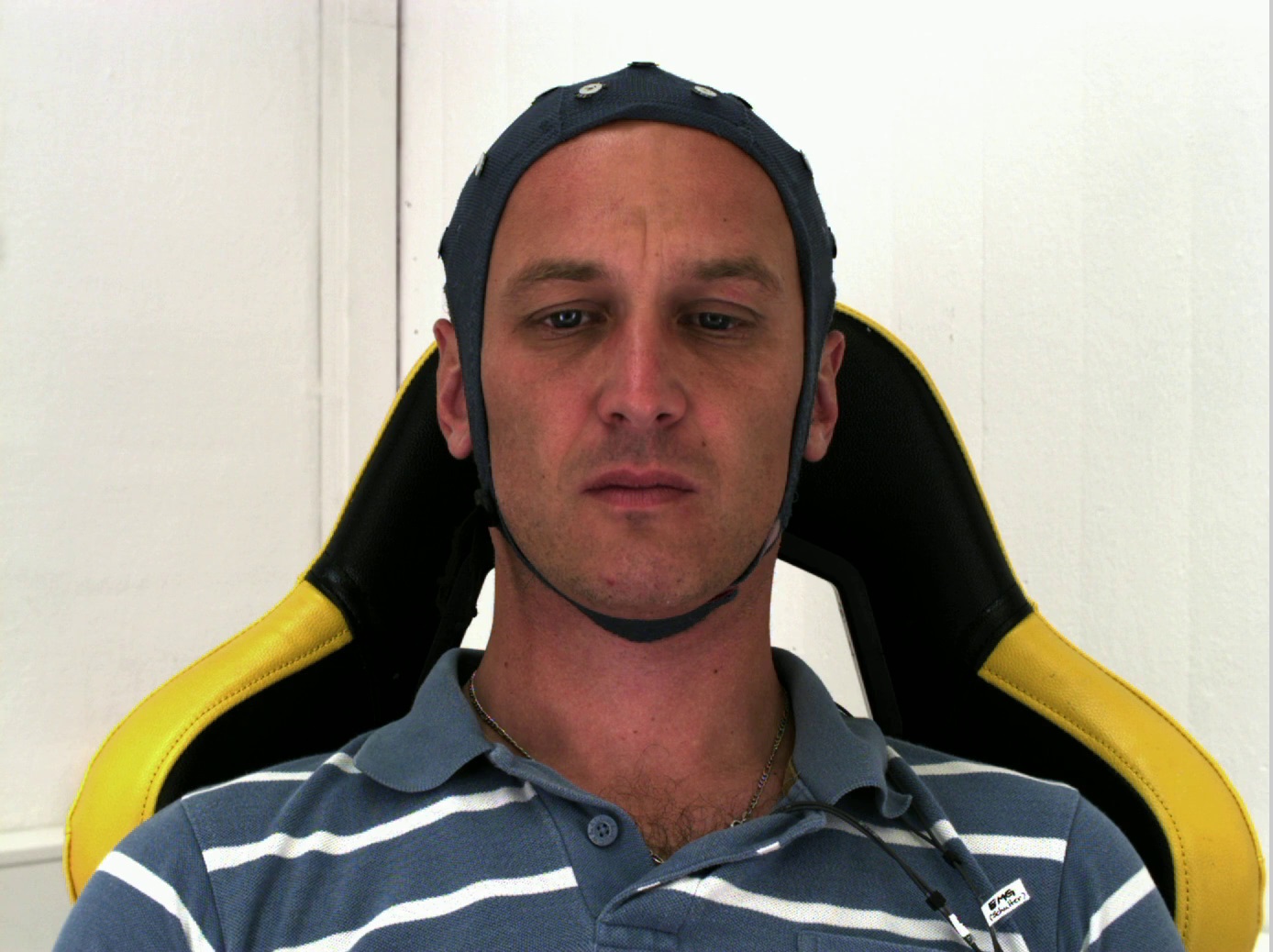}}
~
\subfloat[]{\label{fig:biovid-frame-landmarks}\includegraphics[width=0.32\linewidth]{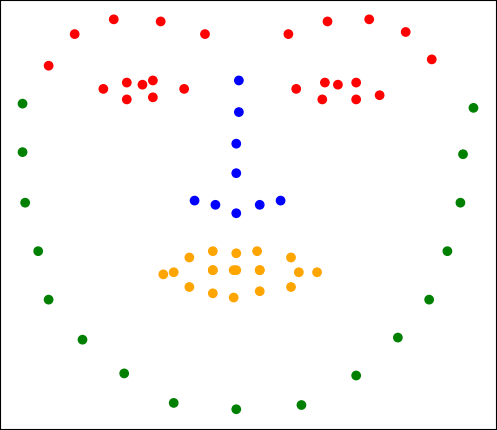}}
~ \\
\subfloat[]{\label{fig:biovid-frame-}\includegraphics[width=0.35\linewidth]{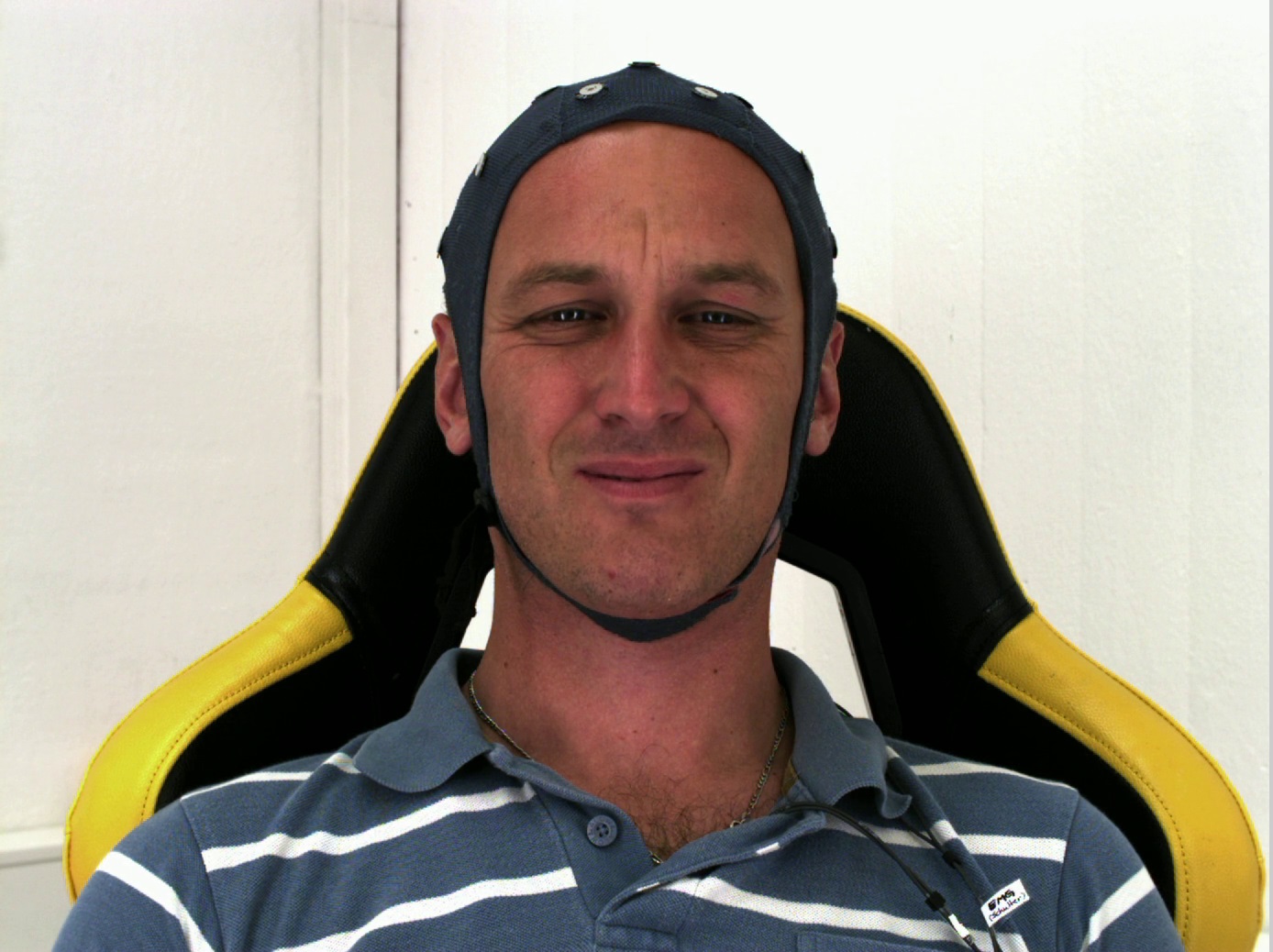}}
~
\subfloat[]{\label{fig:biovid-frame--landmarks}\includegraphics[width=0.32\linewidth]{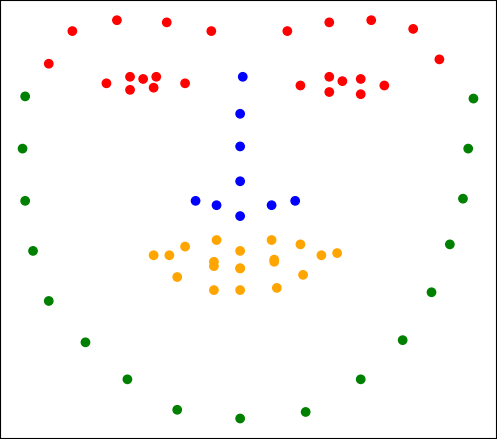}}
\caption{Biovid Heat Pain dataset~\cite{Walter2013Biovid}: Sample images are shown in (a) and (c). In (b) and (d) their corresponding landmark coordinates are evidenced using a different color for each region (best viewed in color).} 
\label{fig:example-frames-biovid}
\end{figure}

This dataset only contains videos and annotations, so we used the OpenPose framework~\cite{cao2017realtime} to extract 70 facial landmarks from each frame of each video in the dataset. The main difference between the landmarks extracted with OpenPose and those distributed with the UNBC-McMaster Shoulder Pain Archive is the addition of 4 landmarks (2 at the extremity of the mouth and 2 at the center of the eyes). Moreover, the landmarks that come with the UNBC-McMaster dataset are extracted using an Active Appearance Model (AAM), that is a semi-automatic algorithm with human in the loop annotation, compared to the fully automatic algorithm proposed by OpenPose. Figure~\ref{fig:example-frames-biovid} shows two frames from a sequence of the Biovid Heat Pain dataset with their corresponding extracted facial landmarks. 

\subsubsection{Results}
The goal here is to estimate the self-reported pain level 
for each sequence of the dataset. The results of our method are obtained using the same two protocols described in the previous section: the LOSO protocol and a 3-fold cross validation. 
The results are summarized in Table~\ref{tab:comparison-results-biovid}. 
For each of these protocols, the curve fitting parameter $\lambda$ and the sampling of each sequence are the same we used on the UNBC-McMaster Shoulder Pain Archive. This means that lambda is equal to $100$ and the sampling rate is set to $50\%$, by taking out one frame every two consecutive frames. Since we observed that face decomposition leads to better results, we only report here our results using the early and late fusion methods, described in Section~\ref{sect:pain-estimation}.

\begin{table}[!ht]
    \centering
    \caption{Biovid Heat Pain dataset: Comparison of results of the proposed method.} 
    \label{tab:comparison-results-biovid}
    \small
    \begin{adjustbox}{width=1\linewidth}
        \begin{tabular}{l|c|c|c}
        	\textbf{Protocol} & \textbf{Regression Setup} & \textbf{MAE} & \textbf{RMSE} \\ 
        	\hline \hline
        	\multirow{2}{*}{LOSO cross validation} & Late fusion & 1.13 & 1.47 \\
        	\cline{2-4} & Early fusion & 1.51 & 1.89 \\ \hline \hline
        	\multirow{2}{*}{3-fold cross validation} & Late fusion & 1.06 & 1.36 \\
        	\cline{2-4} & Early fusion & 1.88 & 2.27 \\ \Xhline{2\arrayrulewidth}
        \end{tabular}
    \end{adjustbox}
\end{table}

\begin{figure}[!ht]
\centering
\includegraphics[width=0.55\linewidth]{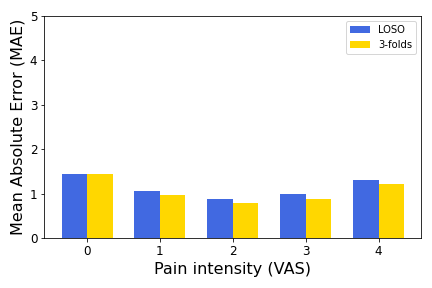}
\caption{MAE per intensity on the Biovid Heat Pain dataset.}
\label{fig:mae_intensity_biovid}
\end{figure}

Using the LOSO cross validation protocol, we obtained a MAE of $1.13$, while we got a MAE of $1.06$ using the 3-fold cross validation protocol with the late fusion strategy. In the same way as with the UNBC-McMaster dataset, we observe an improvement of the results using the late fusion strategy over the early fusion, showing the effectiveness facial decomposition and training of one SVR per region. The overall MAE is lower for the Biovid dataset as there are only 5 different levels of pain, compared to 11 for the UNBC-McMaster dataset and the dataset is larger, meaning that at each round of the cross validation, there are more training data.  
Figure~\ref{fig:mae_intensity_biovid} shows the MAE per intensity obtained using the late fusion strategy and both protocols (\ie, blue bars correspond to the LOSO protocol and yellow bars to the 3-folds cross validation protocol).


\subsubsection{Comparison with state-of-the-art}
As mentioned, the goal of our proposed method is to estimate the pain scores at sequence level for each video of the Biovid Heat Pain dataset. From~\cite{Werner:2019}, most of previous works using this dataset considered the pain estimation as a binary classification problem (presence of pain vs. different intensities of pain) or classified pain intensity in binary pairs. However, some of the approaches focused on continuous pain estimation at sequence level and we report their results in Table~\ref{tab:comparison-sota-biovid}.

\begin{table*}[!hb]
    \centering
    \caption{Comparison of our method with state-of-the-art approaches on the Biovid Heat Pain dataset}
    \label{tab:comparison-sota-biovid}
    \begin{adjustbox}{width=0.8\linewidth}
        \begin{tabular}{l|c|c|c|c|c}
        	\textbf{Method} & \textbf{Protocol} & \textbf{Modalities} & \textbf{Training Labels} & \textbf{MAE} & \textbf{RMSE} \\ \hline\hline
        	Pouromran~\etal~\cite{Pouromran2021BiovidPain} & LOSO & Skin Conductance & VAS & 0.93 & 1.16 \\ \hline
        	Lopez-Martinez and Picard~\cite{LopezMartinez2018PainBiovid} & LOSO & Skin Conductance & VAS & 1.05 & 1.29 \\ \hline
        	Kächele~\etal~\cite{Kachele2017Pain} & LOSO & Bio-signals and videos & VAS & 0.99 & 1.16 \\ \hline
        	Kächele~\etal~\cite{Kachele2017Pain} & LOSO & Statistical geometric features & VAS & 1.16 & 1.35 \\ \hline
        	\textbf{Proposed} & LOSO & Facial landmark coordinates & VAS & 1.13 & 1.47 \\ \hline
        	\textbf{Proposed} & 3-fold CV & Facial landmark coordinates & VAS & 1.06 & 1.36 \\ \Xhline{2\arrayrulewidth}
        \end{tabular}
    \end{adjustbox}
\end{table*}

Kächele~\etal~\cite{Kachele2017Pain} reported multiple results using different modalities to estimate pain indexes. First, the result using early fusion of multiple physiological signals (skin conductance, ECG and EMG) with video features was reported, with a MAE of $0.99$. They also reported a result using statistical geometric features computed after extracting facial landmarks with OpenFace and obtained a MAE of $1.16$.
Both these results were obtained by applying the LOSO cross validation protocol. For a fair analysis, we compared our results with their second result, as it does not use physiological signals. However, their statistical features were extracted from the landmark coordinates, whereas we only used the landmark coordinates and their velocities. Despite of this, we obtained competitive results with a MAE of $1.13$. 

Pouromran~\etal~\cite{Pouromran2021BiovidPain} obtained a MAE of $0.93$ with the LOSO protocol, but using skin conductance as input features. The advantage of the methods using physiological signals can be observed in the different results. However, they required the adoption of intrusive instruments like sensors on the head or on the hand to record bio-signals. Landmark coordinates can be obtained using a simple camera, with no impact on the privacy of each subject.

Table~\ref{tab:comparison-sota-biovid} shows that our approach achieves state of the art results in terms of MAE among approaches using only visual features. If RMSE is considered, the measured accuracy of our approach decreases more than what is observed for the other approaches. Considering that a characterizing trait of the RMSE compared to the MAE is that it gives more relevance to large error values, a plausible interpretation of this pattern is that with our approach there is a residual number of predictions with a large error, yet this error being very low in most of the cases. These predictions with large errors are less frequent in~\cite{Kachele2017Pain} although in most cases the error is higher compared to our approach.

\section{Discussion and Conclusions}\label{sect:conclusions}
We proposed a model for predicting the level of pain based on the dynamics of facial landmarks. The model is based on the decomposition of facial landmarks in different regions of the face and representation of the motion dynamics of these landmarks as trajectories on the Riemannian manifold of fixed rank symmetric positive semi-definite matrices. We have demonstrated the effectiveness of our approach through extensive experiments on the UNBC-McMaster Shoulder Pain Archive and the Biovid dataset. Our approach is competitive with the state-of-the-art on the UNBC-McMaster Shoulder Pain Archive among the approaches that predict the VAS pain score based only on the shape of the face at sequence level. 

The main issue with the proposed method is the time required to compute the kernel of the SVR model. As the size of the dataset increases, the time to compute the similarity matrix used kernel increases as well. 
Future work will investigate solutions to speed-up this computation, for example by clustering the training sequences so as to reduce the number of sequences used to build the kernel. One solution could be based on computing a mean trajectory to represent each pain level index, thus reducing the size of the similarity matrix to compute.
Finally, we also plan to learn the weights for the late fusion strategy, allowing us to better understand the contribution of each region of the face for pain assessment as this remains an open question. This could be addressed through the adoption of a more effective strategy than the grid search approach currently adopted.


\section*{Acknowledgements}
Zakia Hammal’s effort in this publication was supported by
the National Institute Of Nursing Research of the National
Institutes of Health under Awards Number R21NR016510
and R01NR018451. The content is solely the responsibility
of the authors and does not necessarily represent the official
views of the National Institutes of Health. We thank Prof. J-C. Alvarez Paiva from University of Lille for fruitful discussions on the formulation of the distance between $n \times 2$ landmark configurations in~\eqref{eq:distp2}. 
We thank Dr. Pierre-Yves Gousenbourger from Université Catholique de Louvain for providing us the curve fitting MatLab code.

%
%
%
%





\begin{IEEEbiography}[{\includegraphics[width=1in,height=1.25in,clip,keepaspectratio]{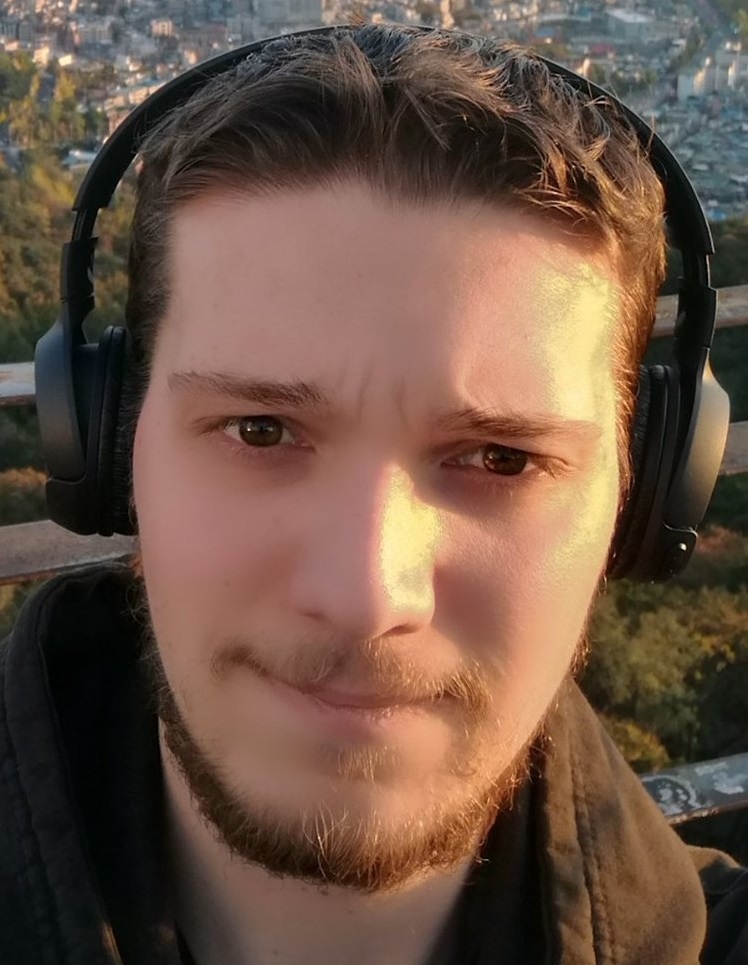}}]{Benjamin Szczapa} is a PhD candidate at the University of Lille and University of Florence. His research interests include computer vision and shape analysis over video sequences.
\end{IEEEbiography}

\begin{IEEEbiography}[{\includegraphics[width=1in,height=1.25in,clip,keepaspectratio]{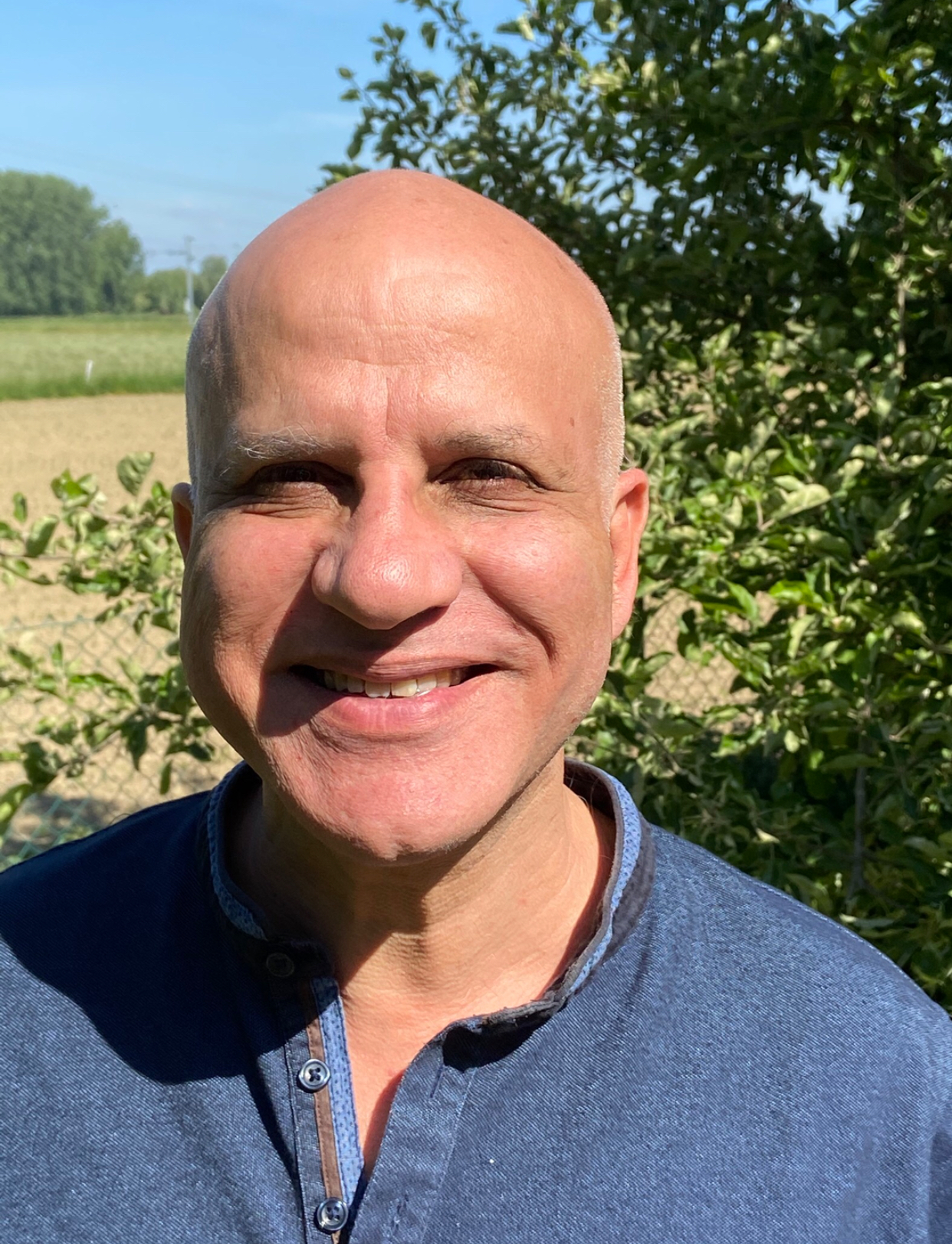}}]{Mohamed Daoudi} is a professor of Computer Science at IMT Nord Europe and a leader of Image group at CRIStAL (UMR CNRS 9189). His research interests include pattern recognition and computer vision and he is the author of over 150 scientific publications. He was the General Chair of IEEE FG 2019. He is/was AE of IEEE Trans. MM, IEEE Trans. On Affective Computing, CVIU, Image and Vision Computing and ACM TOMM. He is a Fellow of IAPR. 
\end{IEEEbiography}

\begin{IEEEbiography}[{\includegraphics[width=1in,height=1.25in,clip,keepaspectratio]{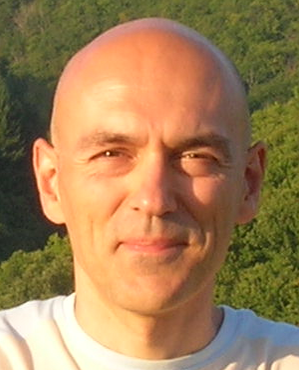}}]{Stefano Berretti} is an Associate Professor at University of Florence. His research interests include 3D computer vision, pattern recognition and multimedia. On these themes, he has published over 200 articles. He is an Associate Editor of the ACM \textit{Trans. on Multimedia Computing, Communications, and Applications}, and an Associate Editor of the IET \textit{Computer Vision journal}. 
\end{IEEEbiography}

\begin{IEEEbiography}[{\includegraphics[width=1in,height=1.25in,clip,keepaspectratio]{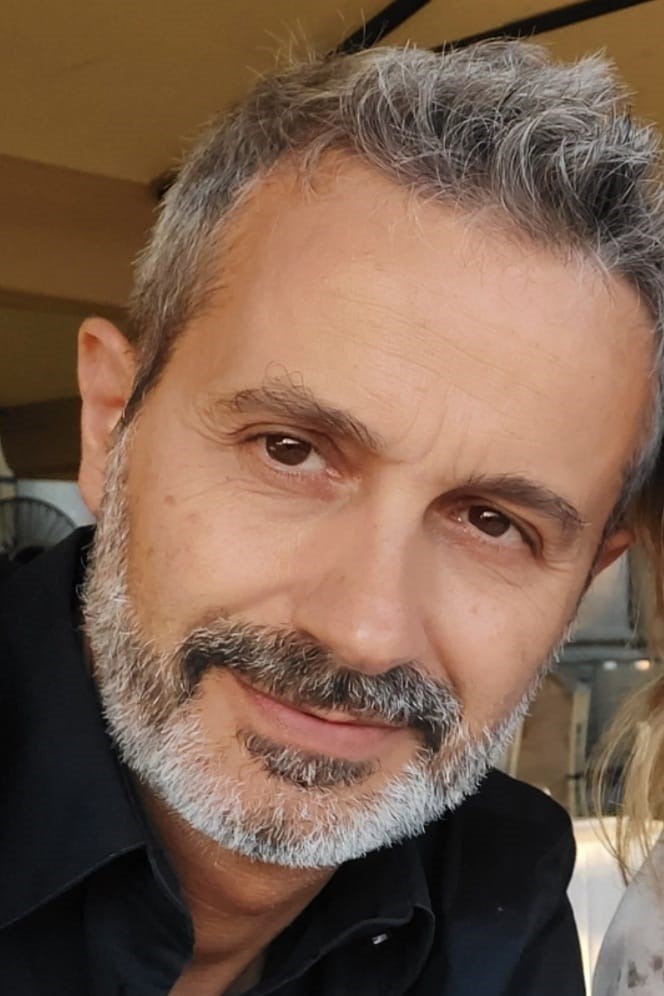}}]{Pietro Pala} received the Ph.D. in Information and Telecommunications Engineering in 1997 at the University of Florence. He is Full Professor of Computer Science and Engineering at the University of Florence. His research activity focuses on 3D data analysis for biometrics and action recognition. He is Associate Editor of the ACM \textit{Trans. on Multimedia Computing, Communications, and Applications}, and of the Springer \textit{Multimedia Systems}. 
\end{IEEEbiography}

\begin{IEEEbiography}[{\includegraphics[width=1in,height=1.25in,clip,keepaspectratio]{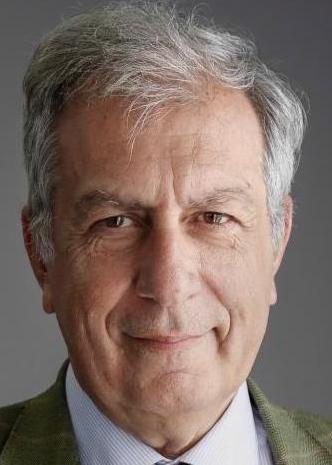}}]{Alberto Del Bimbo} is Full Professor at the University of Firenze. He is the author of over 350 scientific publications in Computer Vision and Multimedia. He was the General Chair of ICPR 2020, ECCV 2012, ACM Multimedia 2010, and IEEE ICMCS 1999, He was the  Editor in Chief of ACM TOMM Transactions on Multimedia Computing, Communications, and Applications.
\end{IEEEbiography}

\begin{IEEEbiography}[{\includegraphics[width=1in,height=1.25in,clip,keepaspectratio]{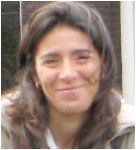}}]{Zakia Hammal}  is a Systems Faculty with the
Robotics Institute Department in the School
of Computer Science at Carnegie Mellon Uni-
versity. Her research overlays the fields of
machine-learning, artificial-intelligence, and so-
cial/behavioral psychology for multimodal human
behavior modeling. She is Associate Editor for
IEEE Transactions on Affective Computing and
IEEE Transactions on Multimedia. She was General Chair of ACM ICMI 2021. Her honors include an Outstanding paper award at ICMI 2012, Best paper award at ACII 2015, and Outstanding
Reviewer Award at FG 2015.
\end{IEEEbiography}

\end{document}